\documentclass[10pt,twocolumn,letterpaper]{article}

\usepackage[pagenumbers]{cvpr} 

\usepackage{booktabs}
\usepackage{caption}
 \usepackage{multirow}
\usepackage[dvipsnames]{xcolor}

\usepackage{amssymb} 
\usepackage{algorithm}
\usepackage{algpseudocode}

\definecolor{cvprblue}{rgb}{0.21,0.49,0.74}
\usepackage[pagebackref,breaklinks,colorlinks,citecolor=green]{hyperref}

\def\paperID{15205} 
\def\confName{CVPR}
\def\confYear{2024}

\title{SAG-ViT: A Scale-Aware, High-Fidelity Patching Approach with Graph Attention for Vision Transformers}

\author{
    Shravan Venkatraman \quad Jaskaran Singh Walia \quad Joe Dhanith P R \\
    Vellore Institute of Technology, Chennai, India \\
    {\tt\small \{shravan.venkatraman18,karanwalia2k3,joe.dhanith\}@gmail.com}
}

\begin{document}
\maketitle
\begin{abstract}
\label{label_abstract}

Vision Transformers (ViTs) have redefined image classification by leveraging self-attention to capture complex patterns and long-range dependencies between image patches. However, a key challenge for ViTs is efficiently incorporating multi-scale feature representations, which is inherent in convolutional neural networks (CNNs) through their hierarchical structure. Graph transformers have made strides in addressing this by leveraging graph-based modeling, but they often lose or insufficiently represent spatial hierarchies, especially since redundant or less relevant areas dilute the image's contextual representation. To bridge this gap, we propose SAG-ViT, a Scale-Aware Graph Attention ViT that integrates multi-scale feature capabilities of CNNs, representational power of ViTs, graph-attended patching to enable richer contextual representation. Using EfficientNetV2 as a backbone, the model extracts multi-scale feature maps, dividing them into patches to preserve richer semantic information compared to directly patching the input images. The patches are structured into a graph using spatial and feature similarities, where a Graph Attention Network (GAT) refines the node embeddings. This refined graph representation is then processed by a Transformer encoder, capturing long-range dependencies and complex interactions. We evaluate SAG-ViT on benchmark datasets across various domains, validating its effectiveness in advancing image classification tasks. Our code and weights are available at \color{pink}\textit{\href{https://github.com/shravan-18/SAG-ViT}{https://github.com/shravan-18/SAG-ViT}}\color{black}.
\end{abstract}    
\section{Introduction}
\label{label_Introduction}

\begin{figure}[t]
  \centering
   \includegraphics[width=\linewidth]{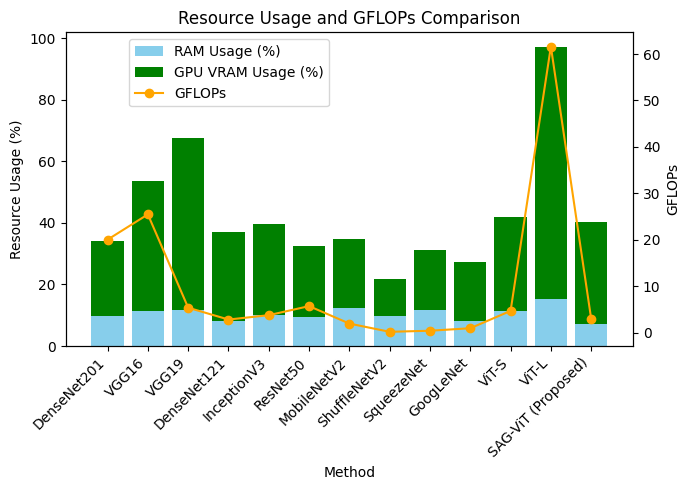}

   \caption{Resource usage and computational complexity comparison across different methods. CNN names refer to a [CNN$\rightarrow$ViT+GAT] architecture. For comparison, ViT-S and ViT-L denote standard Vision Transformer models without the CNN or GAT components.}
   \label{fig:hardwareFig}
\end{figure}

The field of image classification has experienced significant advancements with the introduction of deep learning architectures. CNNs have long been the foundation for image classification tasks due to their proficiency in capturing local spatial hierarchies through convolutional operations \cite{1}. However, their inherent limitations in modeling long-range dependencies restrict their ability to fully exploit global contextual information within images \cite{2}. The introduction of Vision Transformers (ViT) has opened new avenues by leveraging self-attention mechanisms to model global relationships within images. ViTs treat images as sequences of patches (tokens) and have demonstrated competitive performance compared to traditional CNNs \cite{3,4}. Building on these advancements, graph-based transformers have emerged as a promising alternative by representing images as graphs, where nodes capture local features, and edges encode spatial or semantic relationships. This enables better preservation of spatial hierarchies and contextual dependencies, addressing limitations in both CNNs and ViTs by enhancing their ability to process structured representations of image data.

\begin{figure*}[t]
  \centering
   \includegraphics[width=\linewidth]{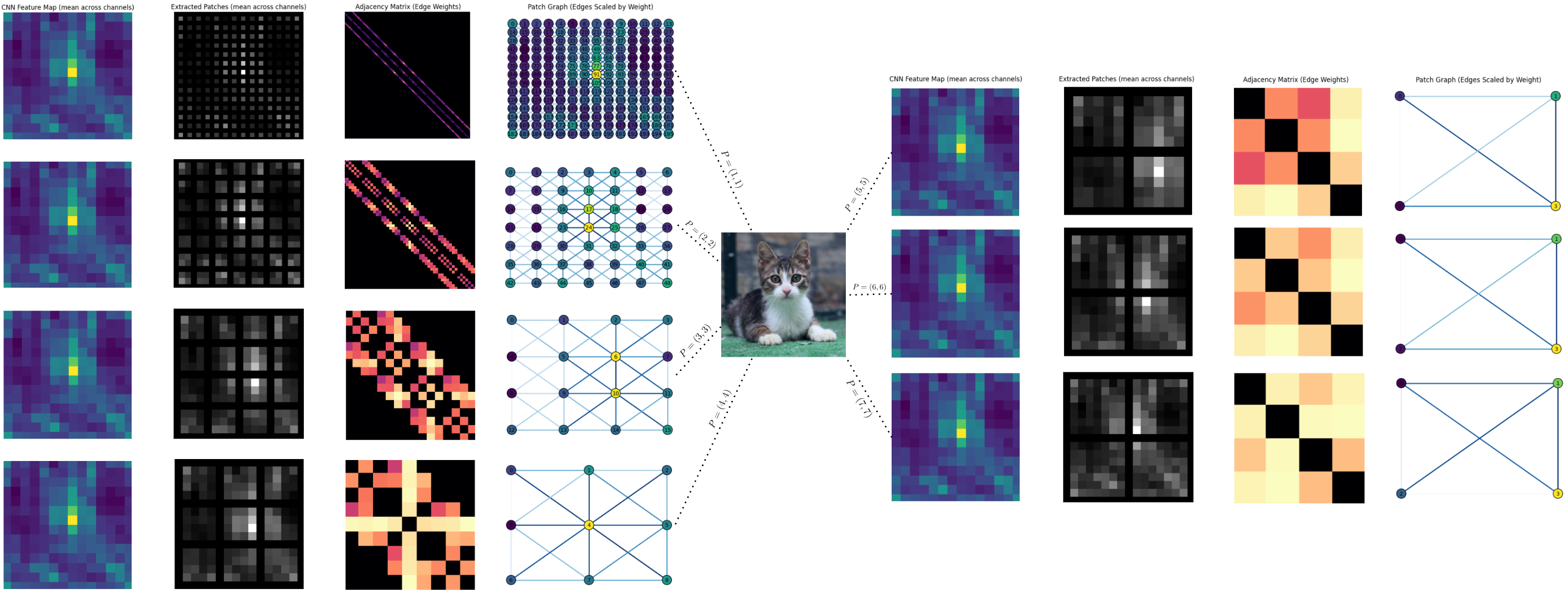}

   \caption{Visualization of the patch generation and graph construction pipeline in SAG-ViT.}
   \label{fig:GraphConstruction}
\end{figure*}

Despite their individual strengths, the above architectures possess inherent limitations, highlighting the need for a more holistic approach to image representation and processing. Recent research has highlighted the importance of multi-scale feature representations in enhancing ViTs’ performance across various vision tasks \cite{6}. Multi-scale approaches enable models to capture objects and patterns of varying sizes, providing a more comprehensive understanding of the image content \cite{fpn}. ViTs often overlook these fine-grained features due to their reliance on directly patching raw images, which not only increases computational overhead but also leads to a loss of semantic context and the rich features provided by intermediate representations. Existing graph-based approaches often construct graphs directly on input images, which lacks hierarchical semantic information, as it bypasses the feature extraction capabilities of convolutional layers. Moreover, constructing graphs at the raw image level often results in redundant or noisy representations, leading to diluted contextual information and increased computational complexity during graph attention operations. 

To address the limitations of existing methods, we propose SAG-ViT, which combines the strengths of CNNs, graph-based modeling, and Transformers. Instead of directly patching raw images, our model extracts rich, multi-scale feature maps using a pre-trained EfficientNetV2 backbone \cite{7}, preserving fine-grained details and semantic context. These feature maps are divided into patches, retaining high-level information and addressing the redundancy and noise issues of graphs constructed directly on raw images. We then construct \textit{k}-connectivity graphs, where nodes represent patches and edges capture spatial and feature relationships, enabling a structured representation of the image context. Using a Graph Attention Network (GAT) \cite{8,9}, we refine node embeddings by dynamically focusing on informative regions, addressing the dilution of contextual information. These enriched embeddings are passed through a Transformer encoder, which integrates local and global features to capture complex relationships for classification.

Our choice of patch size aligns with the receptive field of downstream operations. Smaller patches (\( P \)=(2, 2)) increased computational demands by creating a denser graph, while larger patches  (\( P \)=(6, 6) or more) resulted in fewer nodes but compromised finer semantic details, as smaller features and local patterns were lost during patch aggregation. To balance these trade-offs, we selected a patch size of (4, 4), which ensures effective and efficient graph connectivity, as shown by the adjacency matrix in Figure \ref{fig:GraphConstruction}. Due to this, SAG-ViT achieves significantly lower GPU VRAM and RAM usage compared to the large standalone Transformer (ViT-L) (Figure \ref{fig:hardwareFig}), depicting our feature map-based patching's effectiveness at reducing resource demands.

We validate our proposed SAG-ViT on six diverse benchmark datasets across different domains, including natural images (CIFAR-10) \cite{cifarDataset}, biomedical histopathology (NCT-CRC-HE-100K)  \cite{nctDataset}, agricultural applications (PlantVillage)  \cite{plantvillageDataset}, aerial imagery (NWPU-RESISC45)  \cite{resiscDataset}, traffic sign recognition (GTSRB)  \cite{gtsrbDataset}, and underwater environments (Underwater Trash Dataset) \cite{walia2024deeplearninginnovationsunderwater, utd1}. Our method consistently outperforms existing backbone-assisted transformer-based approaches, demonstrating its capability for efficient and accurate classification. 

Our contributions are summarized as follows:
\begin{itemize} 
\item We propose a novel patching mechanism that operates on CNN-derived feature maps instead of raw images, ensuring the retention of rich semantic information and enabling the efficient capture of multi-scale features. Unlike traditional patching methods, our approach reduces computational complexity while preserving contextual richness. 

\item A unique \textit{k}-connectivity and similarity-based edge weighting scheme is introduced to construct graphs that effectively capture intricate spatial and semantic relationships between patches, addressing the limitations of standard self-attention mechanisms in modeling fine-grained dependencies. 

\item We integrate a Graph Attention Network (GAT) with the Transformer architecture, enabling the effective modeling of both local and global dependencies. We justify this shows a significant improvement over existing backbone-driven Transformer-based methods, which often struggle to balance these relationships. 

\item Our proposed method bridges the gap between multi-scale feature extraction, graph-based representation learning, and Transformers, demonstrating improved and consistent performance in image classification tasks by leveraging hierarchical and relational modeling. 
\end{itemize}

\section{Related Works}
\label{label_related}

\textbf{ViTs for Image Classification } Transformer-based models have gained significant attention in computer vision, initially popularized by the Vision Transformer (ViT), which treats images as sequences of patches and uses self-attention to capture global dependencies, achieving competitive results with CNNs for image classification \cite{9}. However, ViT models often require large datasets and substantial computational resources, limiting their accessibility.

To improve data efficiency, DeiT leverages distillation and data augmentation, enabling ViTs to perform well on smaller datasets \cite{5}. T2T-ViT \cite{11} introduces a Tokens-to-Token transformation to better capture local structures, addressing ViT's limitation of naive tokenization. The Perceiver model uses an asymmetric attention mechanism to distill large inputs into a compact latent space, allowing it to scale effectively for high-dimensional data \cite{12}. Similarly, PVT and CvT incorporate pyramid-like structures into transformers, merging CNN-like multi-scale processing with transformer advantages for richer feature extraction \cite{13}.

The Swin Transformer introduces a shifting window approach to self-attention, efficiently capturing both local and global contexts while maintaining manageable complexity, especially for dense tasks like segmentation and detection \cite{14}. These models highlight a growing trend toward integrating multi-scale representations to improve vision transformers' ability to capture both fine-grained details and long-range dependencies.

\textbf{Multi-Scale Feature Representation} Multi-scale feature representations are critical for recognizing objects and patterns at varying scales \cite{6}. CNNs naturally capture multi-scale features through their hierarchical layers and receptive fields \cite{1}. Techniques such as feature pyramid networks \cite{10} and multi-branch architectures \cite{15} have been proposed to enhance multi-scale learning in CNNs.

In the context of transformers, incorporating multi-scale features remains challenging due to the fixed-size patch tokenization. CrossViT \cite{6} introduces a dual-branch transformer architecture that processes image patches of different sizes in separate branches, fusing them using cross-attention mechanisms. This approach effectively captures both fine-grained details and global context.

\begin{figure*}[t]
  \centering
   \includegraphics[width=\linewidth]{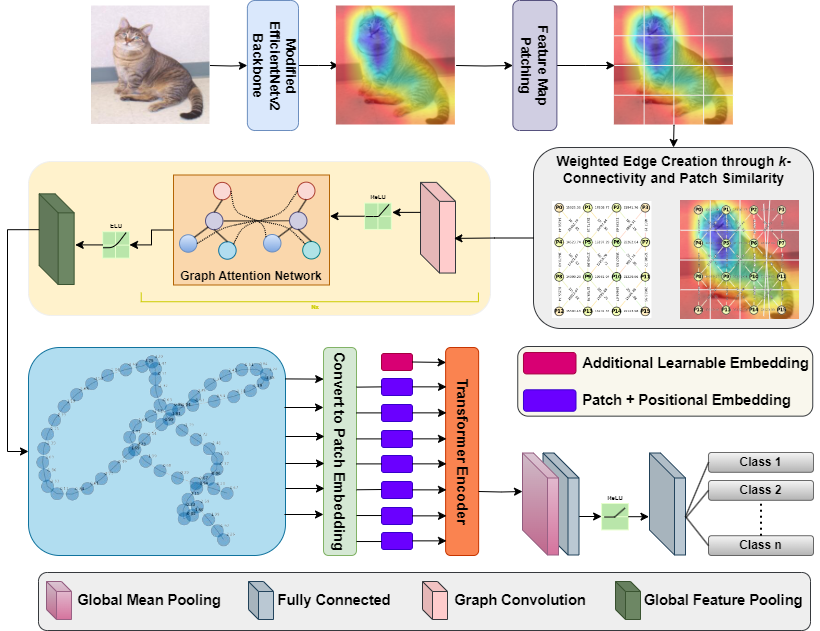}

   \caption{An illustration of our proposed SAG-ViT architecture for learning scale-aware, high-fidelity features with graph attention.}
   \label{fig:architecture}
\end{figure*}

\textbf{Graph Neural Networks for Image Classification } Graph Neural Networks have gained attention for their ability to model relational data. In image classification, representing images as graphs allows for capturing spatial relationships between different regions \cite{9}. Nodes can represent super pixels or patches, and edges encode similarities or spatial connections. Constructing graphs directly from raw images can lead to information loss due to the reduction in spatial resolution \cite{16}. By constructing graphs from CNN-derived feature maps, richer semantic information can be retained \cite{17}. This approach enhances the modeling of complex spatial dependencies crucial for accurate classification.

Graph Attention Networks extend the concept of attention mechanisms to graph-structured data \cite{8}. GATs compute attention coefficients for neighboring nodes, allowing the network to focus on the most relevant connections. This dynamic weighting improves the learning of node representations by emphasizing important relationships. Incorporating GATs in image classification enables the modeling of both local and non-local dependencies \cite{18}. When combined with multi-scale feature representations, GATs can effectively capture intricate patterns within images.

\textbf{Hybrid Models } Recent studies suggest that combining transformer and convolutional layers into a hybrid architecture can harness the strengths of both approaches. BoTNet \cite{19} modifies self-attention in the final three blocks of ResNet to integrate both architectures. The CMT \cite{20} block incorporates depthwise convolutional layers for local feature extraction, alongside a lightweight transformer block. CvT \cite{14} places pointwise and depthwise convolutions before the self-attention mechanism to enhance performance. LeViT \cite{21} replaces the patch embedding block with a convolutional stem, enabling faster inference for image classification. MobileViT \cite{22} combines Transformer blocks with the MobileNetV2 \cite{23} block to create a lightweight vision transformer. Mobile-Former \cite{24} bridges CNNs and transformers in a bidirectional manner to capitalize on both global and local features.

Unlike the above methods, our proposed approach introduces a fundamentally different methodology by unifying multi-scale feature extraction, graph-based relational modeling, and transformer-based long-range dependency learning. To the best of our knowledge, SAG-ViT is the first framework to seamlessly integrate these components, addressing their individual limitations and providing a comprehensive, efficient solution for image classification.

\section{Method: \textit{SAG-ViT}}
\label{label_method}

Figure \ref{fig:architecture} illustrates the network architecture of our proposed Scale-Aware Vision Transformer with Graph Attention (SAG-ViT). We begin by outlining our high-fidelity feature map patching strategy (\S 3.1), which preserves semantic richness while reducing computational overhead. Next, we detail our graph construction methodology (\S 3.2), leveraging \textit{k}-connectivity and feature similarity to capture intricate spatial relationships. Finally, we explain the integration of GAT with Transformer encoders (\S 3.3), enabling the effective modeling of both local and global dependencies for improved classification.

\begin{figure*}[h!]
    \centering
    \begin{subfigure}[b]{0.32\textwidth}
        \centering
        \includegraphics[width=\linewidth]{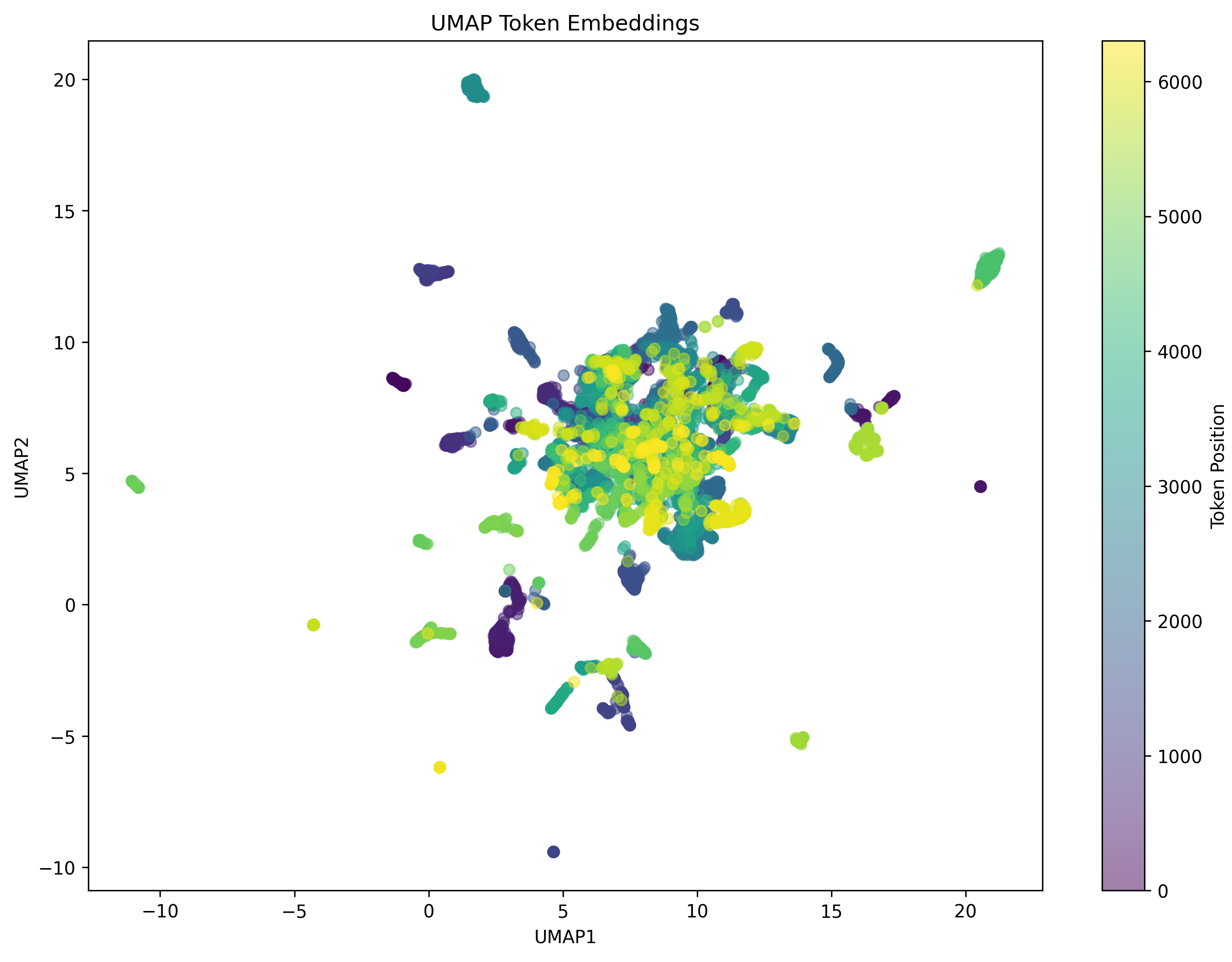} 
        \caption{DeiT}
        \label{fig:deit_umap}
    \end{subfigure}
    \hfill
    \begin{subfigure}[b]{0.32\textwidth}
        \centering
        \includegraphics[width=\linewidth]{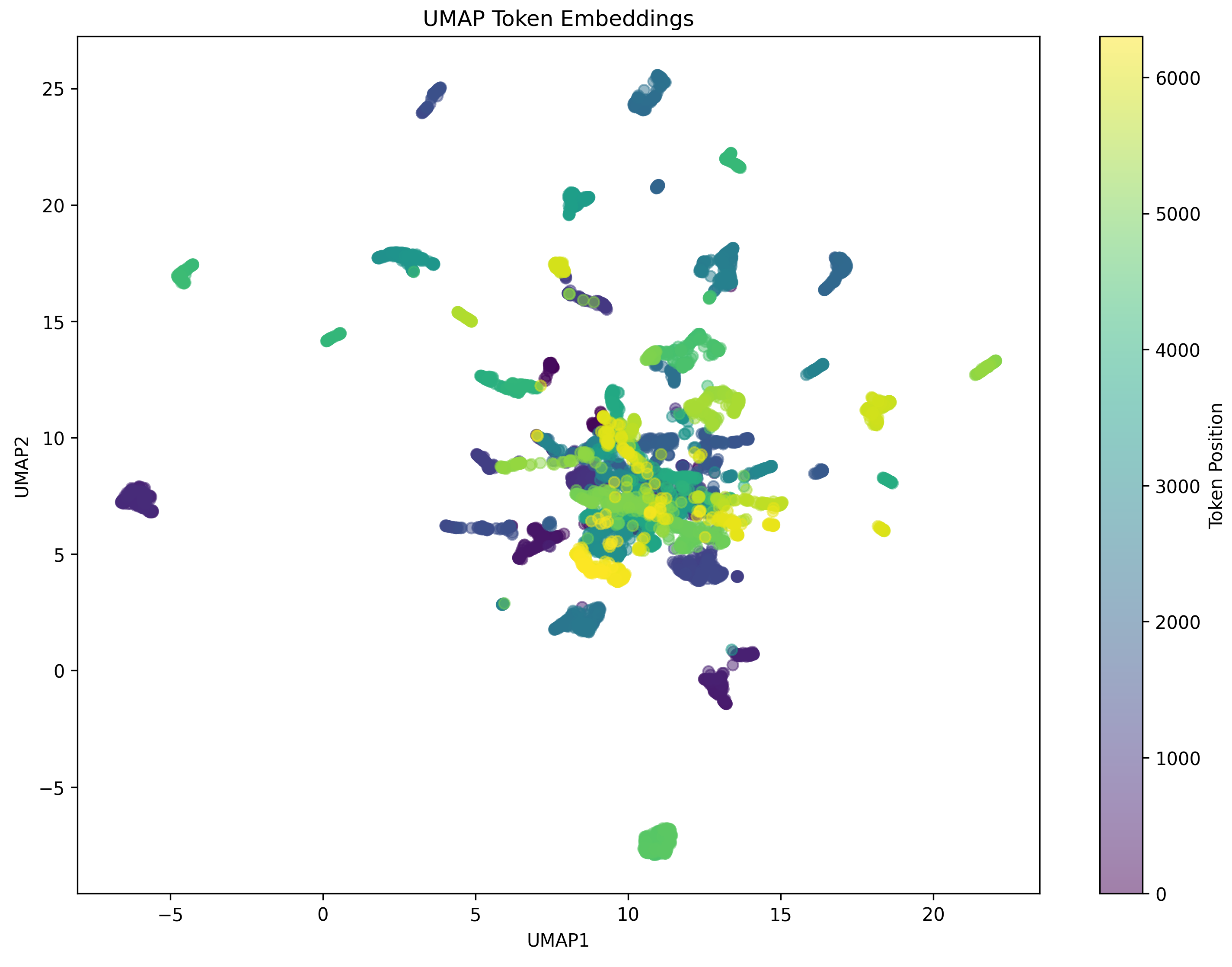} 
        \caption{Vanilla ViT}
        \label{fig:vanillaViT_umap}
    \end{subfigure}
    \hfill
    \begin{subfigure}[b]{0.32\textwidth}
        \centering
        \includegraphics[width=\linewidth]{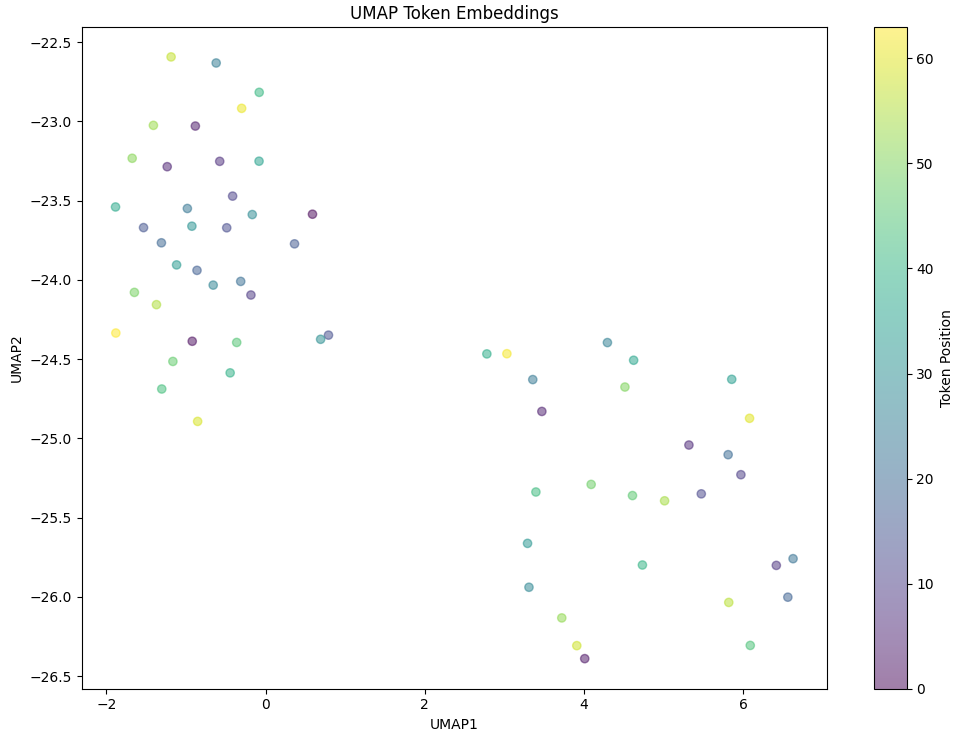} 
        \caption{SAG-ViT (Proposed)}
        \label{fig:sagvit_umap}
    \end{subfigure}
    
    \caption{Feature embeddings projected into a 2D space using UMAP for DeiT, Vanilla ViT, and SAG-ViT (Proposed).}
    \label{fig:umaps}
\end{figure*}

\subsection{High-Fidelity Feature Map Patching}

We initiate the processing pipeline by extracting high-fidelity patches from feature maps generated by a lightweight convolutional backbone. By operating on feature maps rather than raw images, we retain higher-level semantic information. We process the input image $I \in \mathbb{R}^{H \times W \times C}$ through a deep CNN to exploit its compound multiscale feature scaling for receptive fields and efficient convolution paths, yielding a feature map $F \in \mathbb{R}^{H' \times W' \times D}$, where $H' = \frac{H}{s}$, $W' = \frac{W}{s}$, and $D$ denotes the depth of the feature channels with stride $s$. 

To preserve detailed and multi-scale semantic information, we partition the feature map $F$ into non-overlapping patches $P_{i,j} \in \mathbb{R}^{k \times k \times D}$, defined as $P_{i,j} = F[i \cdot k : (i+1) \cdot k,\; j \cdot k : (j+1) \cdot k,\; :]$ for $i \in \{0, \ldots, \frac{H'}{k} - 1\}$ and $j \in \{0, \ldots, \frac{W'}{k} - 1\}$. This operation can be compactly represented using the unfolding operator $\mathcal{U}_k(F)$, where $\mathcal{U}_k : \mathbb{R}^{H' \times W' \times D} \rightarrow \mathbb{R}^{\frac{H'}{k} \times \frac{W'}{k} \times k \times k \times D}$. 

Each patch $P_{i,j}$ is then vectorized into a feature vector $p_{i,j} \in \mathbb{R}^{k^2 D}$, resulting in a collection of patch vectors $\mathcal{P} = \bigcup_{i=0}^{\frac{H'}{k}-1} \bigcup_{j=0}^{\frac{W'}{k}-1} \{p_{i,j}\}$. By extracting patches directly from the feature map $F$, we leverage the high-level abstractions learned by the CNN to ensure that each patch $P_{i,j}$ encapsulates rich semantic information, capturing both local patterns and contextual relationships within the image. Moreover, extracting patches from the reduced spatial dimensions $H' \times W'$ leads to fewer patches $\mathcal{P}$, decreasing computational complexity while maintaining essential information. The vectorized patches $p_{i,j}$ serve as nodes in the subsequent graph construction phase, where their high-dimensional features capture subtle patch relationships, and non-overlapping extraction preserves spatial structure for improved classification.


\subsection{Graph Construction Using \textit{k}-Connectivity and Similarity-Based Edges}

\begin{algorithm}[h!]
\caption{Graph Generation from Image Patches}
\label{alg:patch_graph}
\begin{algorithmic}[1]
\Require Patch location $(r, c)$, Patch dimensions $(p_h, p_w)$, Image size $(H, W)$
\Ensure Constructed Graph $\mathcal{G}$

\Function{GenerateGraph}{$(r, c), (p_h, p_w), (H, W)$}
    \State Initialize graph $\mathcal{G} \gets \varnothing$
    \State Compute grid dimensions: $\text{Rows} \gets H / p_h$, $\text{Cols} \gets W / p_w$
    \State Node identifier: $\nu \gets r \cdot \text{Cols} + c$
    \State $\mathcal{G}.\text{add\_vertex}(\nu)$

    \For{\textbf{each} $(\Delta r, \Delta c) \in \{-1, 0, 1\}^2 \setminus \{(0, 0)\}$}
        \State $\rho \gets r + \Delta r$, $\kappa \gets c + \Delta c$
        \If{$0 \leq \rho < \text{Rows}$ \textbf{and} $0 \leq \kappa < \text{Cols}$}
            \State Neighbor identifier: $\mu \gets \rho \cdot \text{Cols} + \kappa$
            \State $\mathcal{G}.\text{add\_edge}(\nu, \mu)$
        \EndIf
    \EndFor
    \State \Return $\mathcal{G}$
\EndFunction

\end{algorithmic}
\end{algorithm}

Once the patches \(\mathcal{P} = \{p_{i,j}\}\) are extracted, we construct a graph \(G = (V, E)\) to model the spatial and feature-based relationships among them. Here, \(V = \{v_{i,j}\}\) represents the set of nodes corresponding to patches, and \(E\) denotes the set of edges connecting these nodes. Each node \(v_{i,j} \in V\) is associated with a feature vector \(x_{i,j} = p_{i,j} \in \mathbb{R}^{C p^2}\), where each patch of size \((p, p)\) is vectorized into a \(C p^2\)-dimensional feature vector. After extracting all patches, we organize them into a matrix \(X_V = [x_1, x_2, \dots, x_{|\mathcal{V}|}]^T \in \mathbb{R}^{|\mathcal{V}| \times C p^2}\), where \(|\mathcal{V}|\) is the number of patches (nodes) in the graph.

Edges \(e_{u,v} \in E\) are defined based on \(k\)-connectivity and feature similarity. For each patch \(p_i \in V\), its neighboring patches \(p_j \in \mathcal{N}(p_i)\) are spatially adjacent patches within the feature map, determined by a fixed local window size \(k\). The adjacency matrix \(A \in \mathbb{R}^{|\mathcal{V}| \times |\mathcal{V}|}\) is defined as:
\begin{equation}
A_{u,v} = 
\begin{cases} 
\exp \left( -\dfrac{\| x_u - x_v \|_2^2}{\sigma^2} \right) & \text{if } v \in \mathcal{N}_k(u), \\
0 & \text{otherwise},
\end{cases}
\end{equation}
where \(\mathcal{N}_k(u)\) denotes the set of \(k\)-nearest spatial neighbors of node \(u\), and \(\sigma\) is a hyperparameter controlling the decay of the similarity function. The neighborhood function \(\mathcal{N}_k(u)\) is weighted based on the Euclidean distance in the spatial grid, where \(\phi(u)\) maps node \(u\) to its spatial coordinates \((i,j)\), defined as:
\begin{equation}
\begin{aligned}
\mathcal{N}_k(u) = \bigg\{ v \in V \ \bigg| \ &\|\phi(u) - \phi(v)\|_2 \\
&\text{is among the } k \text{ smallest distances} \bigg\}
\end{aligned}
\label{eq:neighborhood_function}
\end{equation}

This graph construction mechanism (Algorithm \ref{alg:patch_graph}) enables the model to capture intricate spatial and feature-based relationships while ensuring computational efficiency through sparsity in the adjacency matrix. 

\begin{figure*}[h]
    \centering
    \resizebox{\textwidth}{!}{%
        \begin{subfigure}[b]{0.3\textwidth}
            \centering
            \includegraphics[width=\textwidth]{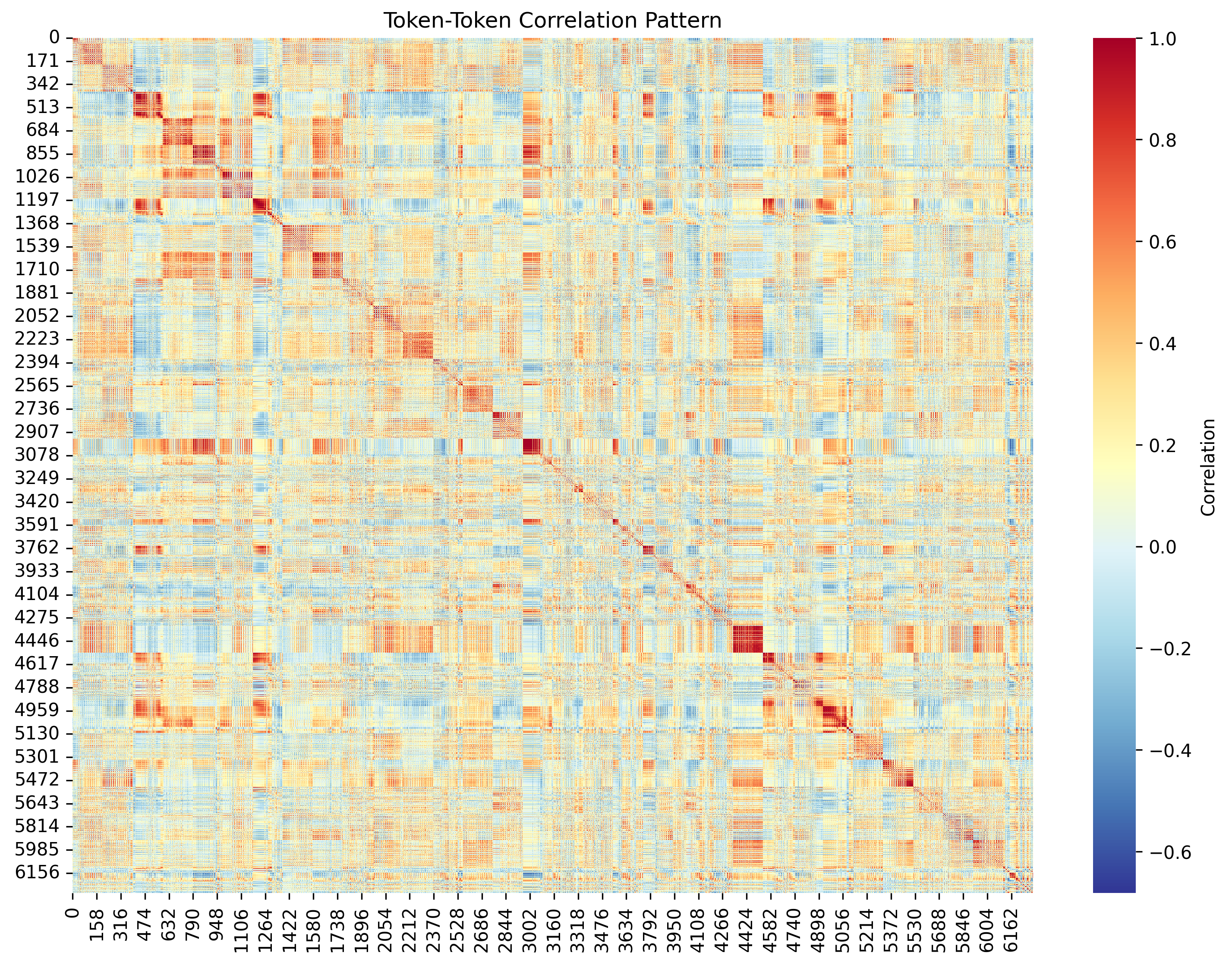} 
            \caption{DeiT}
            \label{fig:deit_corr}
        \end{subfigure}
        \hfill
        \begin{subfigure}[b]{0.3\textwidth}
            \centering
            \includegraphics[width=\textwidth]{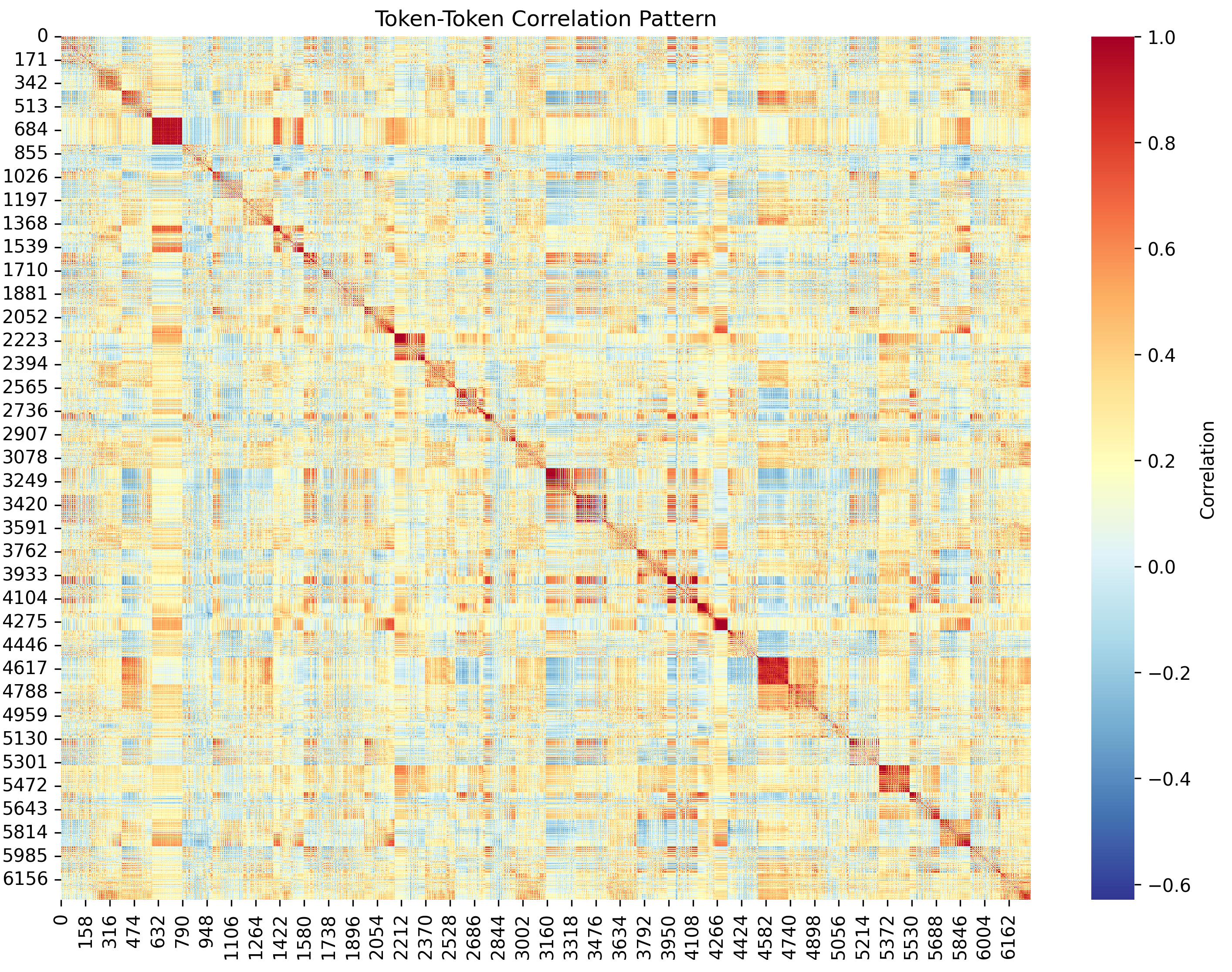} 
            \caption{Vanilla ViT}
            \label{fig:vanillaViT_corr}
        \end{subfigure}
        \hfill
        \begin{subfigure}[b]{0.3\textwidth}
            \centering
            \includegraphics[width=\textwidth]{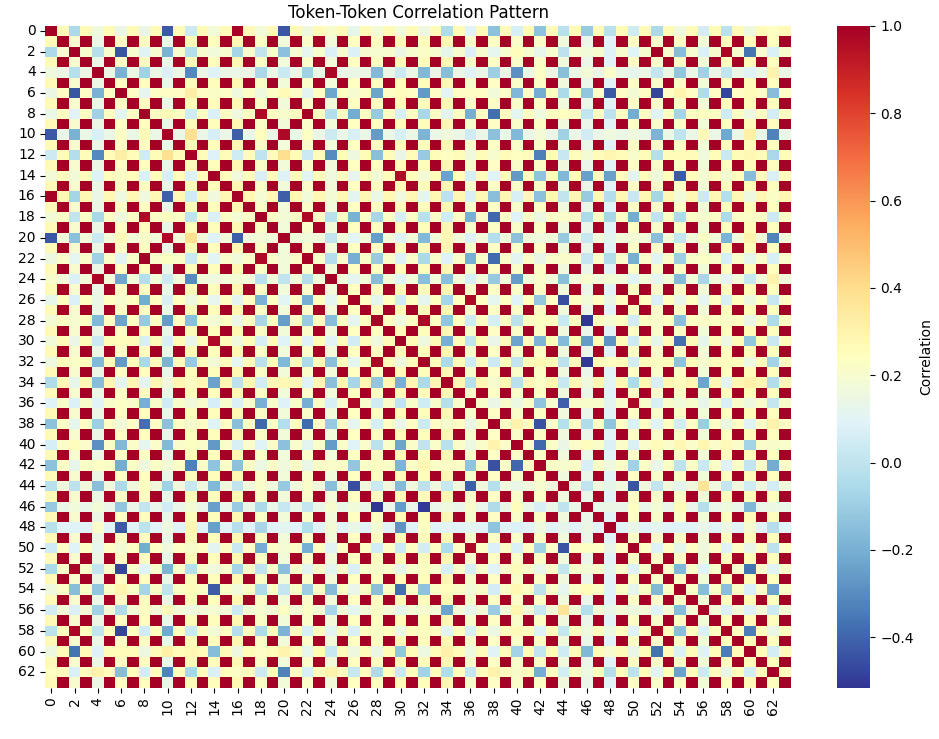} 
            \caption{SAG-ViT (Proposed)}
            \label{fig:sagvit_corr}
        \end{subfigure}
    } 
    \caption{Token-token correlation matrix comparison of SAG-ViT with DeiT and vanilla ViT.}
    \label{fig:ttcs}
\end{figure*}

\subsection{Integration of Graph Attention Networks (GAT) with Transformer Encoders}

After constructing the graph \( G = (V, E) \), we employ a Graph Attention Network (GAT) to process the node features and capture fine-grained dependencies among patches. By integrating GAT with Transformer encoders, we model local and global interactions, improving feature representation (see Algorithm \ref{alg:feature_graph_encoding}). For a given node \( u \), the attention coefficient \( \alpha_{u,v} \) with its neighbor \( v \) is computed as:
\begin{equation}
\alpha_{u,v} = \frac{ \exp\left( \text{LeakyReLU}\left( \mathbf{a}^\top [\mathbf{W} \mathbf{x}_u \| \mathbf{W} \mathbf{x}_v] \right) \right) }{ \sum_{k \in \mathcal{N}(u)} \exp\left( \text{LeakyReLU}\left( \mathbf{a}^\top [\mathbf{W} \mathbf{x}_u \| \mathbf{W} \mathbf{x}_k] \right) \right) },
\end{equation}
where \( \mathbf{W} \in \mathbb{R}^{F' \times D} \) is a learnable transformation matrix, \( \mathbf{a} \in \mathbb{R}^{2F'} \) is a learnable attention vector, and \( \mathcal{N}(u) \) represents the set of neighbors of \( u \). The updated feature \( \mathbf{x}_u' \) for node \( u \) is obtained as \( \mathbf{x}_u' = \text{ReLU}\left( \sum_{v \in \mathcal{N}(u)} \alpha_{u,v} \mathbf{W} \mathbf{x}_v \right) \), where ReLU is a non-linear activation function. For \( h \) attention heads, the concatenated output is defined as:
\begin{equation}
\text{GAT}(\mathbf{x}) = \big\|_{i=1}^h \sigma \left( \sum_{v \in \mathcal{N}(u)} \alpha_{u,v}^{(i)} \mathbf{W}^{(i)} \mathbf{x}_v \right),
\end{equation}
where \( \sigma \) denotes ReLU, and \( \| \) represents concatenation. The resulting node embeddings \( \mathbf{X}' = \{\mathbf{x}_u'\}_{u \in V} \) are fed into a Transformer encoder to model long-range dependencies. Before this, positional encoding \( \mathbf{P} = \{ \mathbf{p}_u \}_{u \in V} \) is added, yielding \( \mathbf{X}'' = \mathbf{X}' + \mathbf{P} \).

The Transformer encoder applies multi-head self-attention using a query vector \( \mathbf{Q} = \mathbf{X}'' \mathbf{W}_Q^{(h)} \), key vector \( \mathbf{K} = \mathbf{X}'' \mathbf{W}_K^{(h)} \), and value vector \( \mathbf{V} = \mathbf{X}'' \mathbf{W}_V^{(h)} \), with:
\begin{equation}
\text{Attention}(\mathbf{Q}, \mathbf{K}, \mathbf{V}) = \text{softmax}\left( \frac{ \mathbf{Q} \mathbf{K}^\top }{ \sqrt{d_k} } \right) \mathbf{V}.
\end{equation}

SAG-ViT captures stronger and more diverse dependencies among patches, as illustrated by the token-token correlation matrix in Figure \ref{fig:ttcs}. This dense and evenly distributed correlation structure demonstrates effective modeling of both local and global interactions. In contrast, DeiT and vanilla ViT exhibit sparser and fragmented correlations, indicating weaker relational modeling.

\begin{algorithm}[h!]
\caption{Feature-Guided Graph Construction and Attention Encoding}
\label{alg:feature_graph_encoding}
\begin{algorithmic}[1]
\Require Feature map $\mathcal{F} \in \mathbb{R}^{B \times C \times H \times W}$, Patch dimensions $(p_h, p_w)$, Transformer parameters $(d_{\text{model}}, n_{\text{heads}})$, GAT parameters $(d_{\text{in}}, d_{\text{hidden}}, d_{\text{out}}, L_{\text{GAT}})$
\Ensure Encoded embeddings $\mathbf{z}_{\text{encoded}}$

\State Partition $\mathcal{F}$ into patches: $\mathcal{P} \in \mathbb{R}^{B \times N \times C \times p_h \times p_w}$, where $N = \frac{H}{p_h} \cdot \frac{W}{p_w}$.
\For{each batch $b \in [1, B]$}
    \State Construct graph $\mathcal{G}_b \gets \bigcup_{n=1}^N \text{PatchToGraph}(\mathcal{P}_n, (p_h, p_w), \mathcal{F})$
\EndFor
\State Aggregate graphs $\{\mathcal{G}_b\}$ for all batches.

\State Compute node features: $\mathbf{X}_b \gets \text{reshape}(\mathcal{P}[b], [N, C \cdot p_h \cdot p_w])$.
\State Assign $\mathcal{G}_b.\mathbf{X} \gets \mathbf{X}_b$ for each $\mathcal{G}_b$.

\State $\mathbf{H}_1 \gets \text{GraphConv}(\mathbf{X}, \mathcal{E}; d_{\text{in}}, d_{\text{hidden}})$
\For{$l = 1, \dots, L_{\text{GAT}} - 1$}
    \State $\mathbf{H}_{l+1} \gets \text{ReLU} \big( \text{GAT}(\mathbf{H}_l, \mathcal{E}) \big)$
\EndFor
\State $\mathbf{H}_{\text{final}} \gets \text{GAT}(\mathbf{H}_{L_{\text{GAT}}}, \mathcal{E}; d_{\text{hidden}}, d_{\text{out}})$
\State $\mathbf{H}_{\text{pooled}} \gets \text{globalMeanPooling}(\mathbf{H}_{\text{final}}, B)$.

\For{$t = 1, \dots, L_{\text{Transformer}}$}
    \State $\mathbf{H}_{t+1} \gets \text{FeedForward} \big( \text{MultiHeadAttn}(\mathbf{H}_t, n_{\text{heads}}) \big)$
\EndFor
\State $\mathbf{z}_{\text{encoded}} \gets \mathbf{H}_{t+1}$.

\State \Return Encoded embeddings $\mathbf{z}_{\text{encoded}}$
\end{algorithmic}
\end{algorithm}

This two-stage process is formalized as \( \mathbf{X}' = \text{GAT}(\mathbf{X}, \mathbf{A}) \) followed by \( \mathbf{X}''' = \text{Transformer}(\mathbf{X}' + \mathbf{P}) \). To aggregate embeddings, a global mean pooling operation is applied:
\begin{equation}
\mathbf{z} = \frac{1}{|V|} \sum_{u \in V} \mathbf{x}_u'''.
\end{equation}
Finally, the pooled representation \( \mathbf{z} \) is passed through a Multi-Layer Perceptron (MLP) to produce classification logits:
\begin{equation}
\hat{\mathbf{Y}} = \text{softmax}(\mathbf{W}_{\text{out}} \mathbf{z} + \mathbf{b}_{\text{out}}),
\end{equation}
where \( \mathbf{W}_{\text{out}} \in \mathbb{R}^{C \times d_{\text{model}}} \) and \( \mathbf{b}_{\text{out}} \in \mathbb{R}^C \) are learnable parameters, and \( C \) is the number of target classes.

To further evaluate the quality of token representations, we visualized UMAP embeddings of token features (Figure \ref{fig:umaps}). Our method exhibits more dispersed embeddings, reflecting greater diversity in token representations. By comparison, DeiT and vanilla ViT show compact but less distinct clusters, implying limited capacity for representing complex feature relationships. This hierarchical processing first refines patch embeddings using graph-based attention to capture localized relationships, followed by Transformer-based self-attention to integrate these embeddings into a cohesive global representation.

\section{Experiments and Results}
\label{label_results}

We conduct experiments and provide a thorough result analysis on six diverse benchmark datasets: CIFAR-10 \cite{cifarDataset}, GTSRB \cite{gtsrbDataset}, NCT-CRC-HE-100K \cite{nctDataset}, NWPU-RESISC45 \cite{resiscDataset}, PlantVillage \cite{plantvillageDataset}, and Underwater Trash Dataset (UTD) \cite{walia2024deeplearninginnovationsunderwater, utd1}. Finally, we present an ablation study on the various components of SAG-ViT. 

\textbf{Training Settings } We primarily adopt the training settings described in \cite{touvron2020training}. For all model variants, we standardized the input image resolution to 224\textsuperscript{2} for consistent evaluation and comparison. We trained the model for 128 epochs using the Adam optimizer \cite{AdamOpt}. A cosine decay learning rate schedule with a 10-epoch linear warm-up phase was employed. Training was conducted with a batch size of 128, an initial learning rate of 0.001, weight decay of 0.01, and gradient clipping with a maximum norm of 1. 

\textbf{Hardware Specifications } The experiments were conducted on a system running Linux 5.15.133, equipped with an AMD EPYC 7763 processor featuring an x86\_64 architecture. The CPU configuration included 128 cores, organized as 64 cores per socket across 2 sockets, with 1 thread per core. The GPU computations were powered by the AMD Instinct™ MI210 accelerator, built on the CDNA2 architecture. It is equipped with 64 GB of HBM2e memory with ECC support, operating on a 4096-bit memory interface at a clock speed of 1.6 GHz, delivering a memory bandwidth of up to 1.6 TB/s. The GPU operates with a base clock speed of 1.0 GHz and can boost up to 1.7 GHz. The maximum power consumption of the GPU is 300 W, supported by a single 8-pin power connector.


\subsection{Overall Performance}

\begin{figure}[h]
  \centering
   \includegraphics[width=\linewidth]{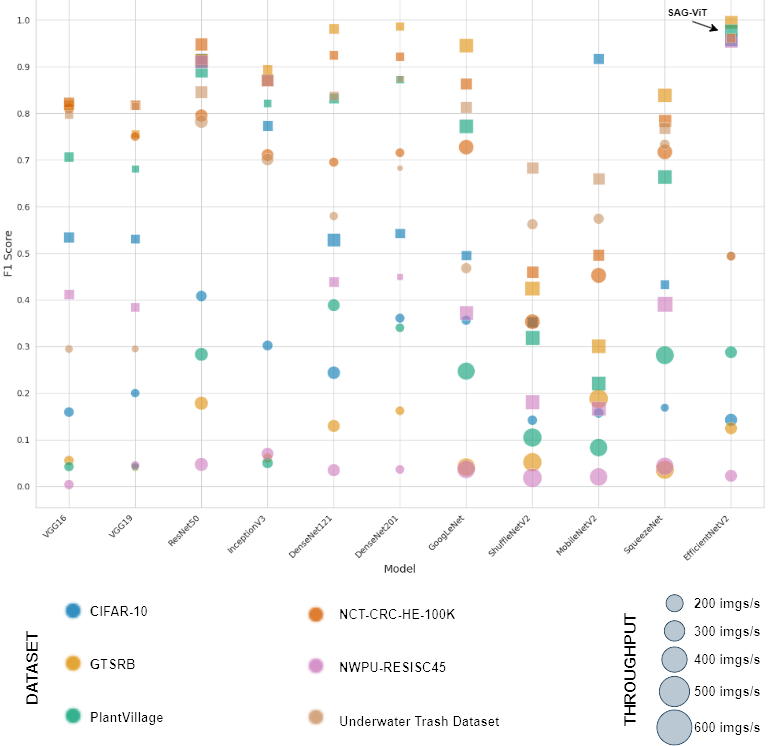}

\caption{F1 Score and Throughput comparison of SAG-ViT and baseline methods across datasets ($\bullet$: without GAT, $\square$: with GAT).}
   \label{fig:QuantCompCustom}
\end{figure}

We present a performance comparison of methods with and without the GAT module in Tables \ref{tab:cifar10} to \ref{tab:utd}. Each method represents a [CNN$\rightarrow$ViT] architecture, where $\text{CNN}_t$ denotes the CNN backbone followed by a ViT module. Rows labeled "+ GAT" indicate the corresponding $\text{CNN}_t$ architecture enhanced with a GAT module. Throughput (\(\mathcal{T}\)) in images per second, F1 scores (\(F_1\)), and percentage improvements (\(\Delta F_1\)) are reported for each dataset. Figure~\ref{fig:QuantCompCustom} compares the F1 scores and throughput of SAG-ViT against baseline methods across multiple datasets.

\begin{table}[h!]
\centering
\caption{Performance comparison on the CIFAR-10 dataset.}
\label{tab:cifar10}
\resizebox{\columnwidth}{!}{
\begin{tabular}{l|lccc}
\hline
\textbf{Model} & \textbf{Method} & \textbf{(\(\mathcal{T}\)) (images/s)} & \textbf{\(F_1\)} & \textbf{\(\Delta F_1\) (\%)} \\ \hline
\multirow{2}{*}{$\text{VGG16 \cite{vgg}}$}          & $\text{VGG16}_t$  & 271.49 & 0.1594 & -       \\ 
                                         & + GAT            & 282.71 & 0.5345 & 235.00  \\ \hline
\multirow{2}{*}{$\text{VGG19 \cite{vgg}}$}          & $\text{VGG19}_t$  & 232.71 & 0.2001 & -       \\ 
                                         & + GAT            & 244.22 & 0.5307 & 165.00  \\ \hline
\multirow{2}{*}{$\text{ResNet50 \cite{resnet}}$}       & $\text{ResNet50}_t$ & 299.22 & 0.4082 & -       \\ 
                                         & + GAT            & 311.34 & 0.9172 & 125.00  \\ \hline
\multirow{2}{*}{$\text{InceptionV3 \cite{inception}}$}    & $\text{InceptionV3}_t$ & 280.32 & 0.3021 & -       \\ 
                                         & + GAT            & 273.23 & 0.7734 & 157.80  \\ \hline
\multirow{2}{*}{$\text{DenseNet121 \cite{densenet}}$}    & $\text{DenseNet121}_t$ & 375.92 & 0.2438 & -       \\ 
                                         & + GAT            & 387.93 & 0.5290 & 117.00  \\ \hline
\multirow{2}{*}{$\text{DenseNet201 \cite{densenet}}$}    & $\text{DenseNet201}_t$ & 250.15 & 0.3607 & -       \\ 
                                         & + GAT            & 267.54 & 0.5427 & 50.50   \\ \hline
\multirow{2}{*}{$\text{GoogLeNet \cite{googlenet}}$}      & $\text{GoogLeNet}_t$ & 255.52 & 0.3562 & -       \\ 
                                         & + GAT            & 266.23 & 0.4954 & 39.00   \\ \hline
\multirow{2}{*}{$\text{ShuffleNetV2 \cite{shufflenet}}$}   & $\text{ShuffleNetV2}_t$ & 264.45 & 0.1419 & -       \\ 
                                         & + GAT            & 288.65 & 0.3523 & 148.20  \\ \hline
\multirow{2}{*}{$\text{MobileNetV2 \cite{23}}$}    & $\text{MobileNetV2}_t$ & 277.98 & 0.1577 & -       \\ 
                                         & + GAT            & 285.18 & 0.9169 & 482.00  \\ \hline
\multirow{2}{*}{$\text{SqueezeNet \cite{squeezenet}}$}     & $\text{SqueezeNet}_t$ & 221.29 & 0.1688 & -       \\ 
                                         & + GAT            & 232.37 & 0.4328 & 156.00  \\ \hline
\multirow{2}{*}{$\text{EfficientNetV2 \cite{effnet}}$} & $\text{EfficientNetV2}_t$ & 362.78 & 0.1428 & -       \\ 
                                         & + GAT (Proposed SAG-ViT) & 372.98 & 0.9574 & 570.00  \\ \hline
\end{tabular}
}
\end{table}

\begin{table}[h!]
\centering
\caption{Performance comparison on the GTSRB dataset.}
\label{tab:gtsrb}
\resizebox{\columnwidth}{!}{
\begin{tabular}{l|lccc}
\hline
\textbf{Model} & \textbf{Method} & \textbf{(\(\mathcal{T}\)) (images/s)} & \textbf{\(F_1\)} & \textbf{\(\Delta F_1\) (\%)} \\ \hline
\multirow{2}{*}{$\text{VGG16 \cite{vgg}}$} 
    & $\text{VGG16}_t$ & 265.83 & 0.0555 & -       \\ 
    & + GAT            & 278.91 & 0.8180 & 1369.10 \\ \hline
\multirow{2}{*}{$\text{VGG19 \cite{vgg}}$} 
    & $\text{VGG19}_t$ & 213.13 & 0.0411 & -       \\ 
    & + GAT            & 221.41 & 0.7551 & 1761.10 \\ \hline
\multirow{2}{*}{$\text{ResNet50 \cite{resnet}}$} 
    & $\text{ResNet50}_t$ & 405.67 & 0.1783 & -       \\ 
    & + GAT            & 395.73 & 0.9134 & 412.10  \\ \hline
\multirow{2}{*}{$\text{InceptionV3 \cite{inception}}$} 
    & $\text{InceptionV3}_t$ & 250.34 & 0.0612 & -       \\ 
    & + GAT            & 260.83 & 0.8934 & 1489.00 \\ \hline
\multirow{2}{*}{$\text{DenseNet121 \cite{densenet}}$} 
    & $\text{DenseNet121}_t$ & 358.63 & 0.1296 & -       \\ 
    & + GAT            & 265.58 & 0.9813 & 657.30  \\ \hline
\multirow{2}{*}{$\text{DenseNet201 \cite{densenet}}$} 
    & $\text{DenseNet201}_t$ & 242.95 & 0.1623 & -       \\ 
    & + GAT            & 224.74 & 0.9862 & 508.70  \\ \hline
\multirow{2}{*}{$\text{GoogLeNet \cite{googlenet}}$} 
    & $\text{GoogLeNet}_t$ & 636.68 & 0.0413 & -       \\ 
    & + GAT            & 419.01 & 0.9455 & 2170.00 \\ \hline
\multirow{2}{*}{$\text{ShuffleNetV2 \cite{shufflenet}}$} 
    & $\text{ShuffleNetV2}_t$ & 674.13 & 0.0519 & -       \\ 
    & + GAT            & 453.13 & 0.4244 & 717.00  \\ \hline
\multirow{2}{*}{$\text{MobileNetV2 \cite{23}}$} 
    & $\text{MobileNetV2}_t$ & 697.16 & 0.1880 & -       \\ 
    & + GAT            & 422.09 & 0.3006 & 59.60   \\ \hline
\multirow{2}{*}{$\text{SqueezeNet \cite{squeezenet}}$} 
    & $\text{SqueezeNet}_t$ & 659.54 & 0.0357 & -       \\ 
    & + GAT            & 421.12 & 0.8392 & 2206.30 \\ \hline
\multirow{2}{*}{$\text{EfficientNetV2 \cite{effnet}}$} 
    & $\text{EfficientNetV2}_t$ & 354.12 & 0.1246 & -       \\ 
    & + GAT (Proposed SAG-ViT) & 371.87 & 0.9958 & 696.50  \\ \hline
\end{tabular}
}
\end{table}

\begin{table}[h!]
\centering
\caption{Performance comparison on the PlantVillage dataset.}
\label{tab:plantvillage}
\resizebox{\columnwidth}{!}{
\begin{tabular}{l|lccc}
\hline
\textbf{Model} & \textbf{Method} & \textbf{(\(\mathcal{T}\)) (images/s)} & \textbf{\(F_1\)} & \textbf{\(\Delta F_1\) (\%)} \\ \hline
\multirow{2}{*}{$\text{VGG16 \cite{vgg}}$} 
    & $\text{VGG16}_t$ & 260.27 & 0.0424 & -       \\ 
    & + GAT            & 256.79 & 0.7064 & 1669.00 \\ \hline
\multirow{2}{*}{$\text{VGG19 \cite{vgg}}$} 
    & $\text{VGG19}_t$ & 207.82 & 0.0424 & -       \\ 
    & + GAT            & 197.69 & 0.6811 & 1502.00 \\ \hline
\multirow{2}{*}{$\text{ResNet50 \cite{resnet}}$} 
    & $\text{ResNet50}_t$ & 397.12 & 0.2832 & -       \\ 
    & + GAT            & 370.93 & 0.8905 & 214.00  \\ \hline
\multirow{2}{*}{$\text{InceptionV3 \cite{inception}}$} 
    & $\text{InceptionV3}_t$ & 300.00 & 0.0503 & -       \\ 
    & + GAT            & 210.00 & 0.8216 & 1543.00 \\ \hline
\multirow{2}{*}{$\text{DenseNet121 \cite{densenet}}$} 
    & $\text{DenseNet121}_t$ & 355.62 & 0.3886 & -       \\ 
    & + GAT            & 269.56 & 0.8321 & 114.00  \\ \hline
\multirow{2}{*}{$\text{DenseNet201 \cite{densenet}}$} 
    & $\text{DenseNet201}_t$ & 241.13 & 0.3401 & -       \\ 
    & + GAT            & 216.09 & 0.8725 & 156.00  \\ \hline
\multirow{2}{*}{$\text{GoogLeNet \cite{googlenet}}$} 
    & $\text{GoogLeNet}_t$ & 603.65 & 0.2471 & -       \\ 
    & + GAT            & 419.92 & 0.7726 & 214.00  \\ \hline
\multirow{2}{*}{$\text{ShuffleNetV2 \cite{shufflenet}}$} 
    & $\text{ShuffleNetV2}_t$ & 675.26 & 0.1048 & -       \\ 
    & + GAT            & 441.22 & 0.3190 & 205.00  \\ \hline
\multirow{2}{*}{$\text{MobileNetV2 \cite{23}}$} 
    & $\text{MobileNetV2}_t$ & 623.44 & 0.0832 & -       \\ 
    & + GAT            & 423.92 & 0.2213 & 165.00  \\ \hline
\multirow{2}{*}{$\text{SqueezeNet \cite{squeezenet}}$} 
    & $\text{SqueezeNet}_t$ & 649.82 & 0.2813 & -       \\ 
    & + GAT            & 423.71 & 0.6638 & 136.00  \\ \hline
\multirow{2}{*}{$\text{EfficientNetV2 \cite{effnet}}$} 
    & $\text{EfficientNetV2}_t$ & 348.66 & 0.2876 & -       \\ 
    & + GAT (Proposed SAG-ViT) & 371.92 & 0.9772 & 240.00  \\ \hline
\end{tabular}
}
\end{table}

\textbf{CIFAR-10 } Table~\ref{tab:cifar10} compares the proposed SAG-ViT against other models on CIFAR-10. EfficientNetV2$\rightarrow$ViT yields a baseline F1 score of 0.1428, which improves to 0.9574 (+570.00\%) with the addition of GAT in SAG-ViT. SAG-ViT also outperforms MobileNetV2$\rightarrow$ViT + GAT (F1 = 0.9169) and DenseNet121$\rightarrow$ViT + GAT (F1 = 0.5290) by +4.41\% and +80.95\%, respectively. In terms of throughput, SAG-ViT processes 372.98 images/s, maintaining efficiency compared to other GAT-enhanced models such as MobileNetV2 (285.18 images/s) and DenseNet121 (387.93 images/s), indicating a strong balance between computational cost and accuracy.

\begin{figure*}[h!]
\centering
\begin{subfigure}[b]{0.3\textwidth}
    \centering
    \includegraphics[width=\textwidth]{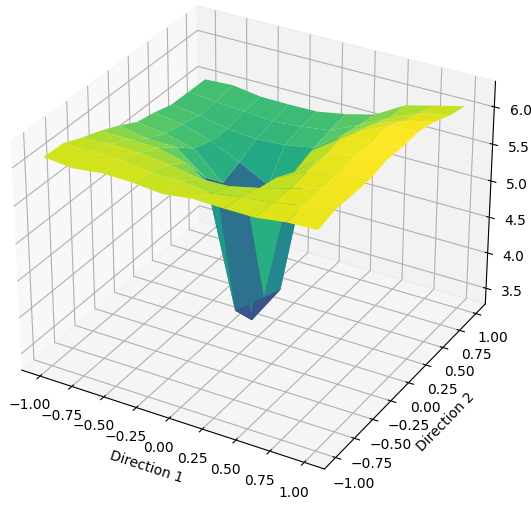}
    \caption{Loss landscape for the convolutional layer.}
    \label{fig:landsape1}
\end{subfigure}
\hfill
\begin{subfigure}[b]{0.3\textwidth}
    \centering
    \includegraphics[width=\textwidth]{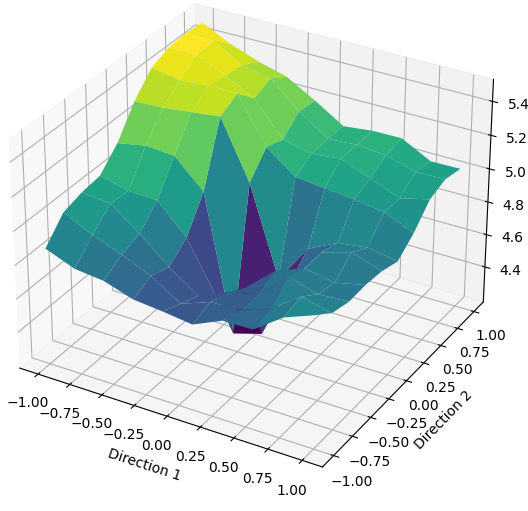}
    \caption{Loss landscape for the graph layer.}
    \label{fig:landsape2}
\end{subfigure}
\hfill
\begin{subfigure}[b]{0.3\textwidth}
    \centering
    \includegraphics[width=\textwidth]{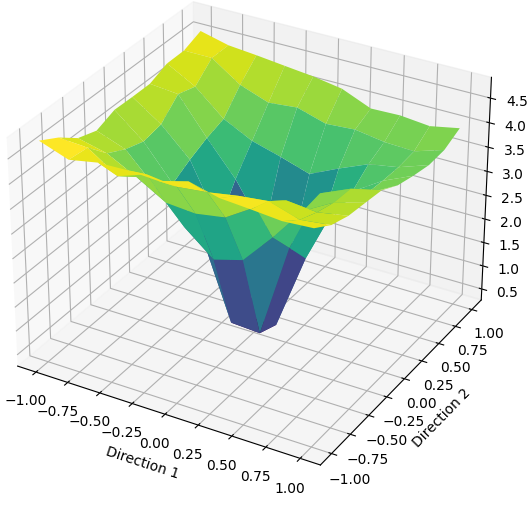}
    \caption{Loss landscape for the Transformer layer.}
    \label{fig:landsape3}
\end{subfigure}

\caption{Comparison of loss landscapes for the final convolutional, graph attention, and Transformer layers, showcasing parameter sensitivity to perturbations.}
\label{fig:landscapes}
\end{figure*}

\textbf{GTSRB } On GTSRB (Table~\ref{tab:gtsrb}), SAG-ViT achieves the highest F1 score of 0.9958, representing a +696.50\% improvement over EfficientNetV2$\rightarrow$ViT and surpassing DenseNet121$\rightarrow$ViT + GAT (F1 = 0.9813) by +1.49\%. SAG-ViT also demonstrates competitive throughput (371.87 images/s), slightly lower than lightweight models like MobileNetV2$\rightarrow$ViT + GAT (422.09 images/s) but significantly higher than DenseNet201$\rightarrow$ViT + GAT (224.74 images/s). 

\begin{table}[t!]
\centering
\caption{Performance comparison on the NCT-CRC-HE-100K dataset.}
\label{tab:nctcrc}
\resizebox{\columnwidth}{!}{
\begin{tabular}{l|lccc}
\hline
\textbf{Model} & \textbf{Method} & \textbf{(\(\mathcal{T}\)) (images/s)} & \textbf{\(F_1\)} & \textbf{\(\Delta F_1\) (\%)} \\ \hline
\multirow{2}{*}{$\text{VGG16 \cite{vgg}}$} 
    & $\text{VGG16}_t$ & 281.81 & 0.8108 & -       \\ 
    & + GAT            & 291.32 & 0.8234 & 1.60    \\ \hline
\multirow{2}{*}{$\text{VGG19 \cite{vgg}}$} 
    & $\text{VGG19}_t$ & 247.34 & 0.7505 & -       \\ 
    & + GAT            & 267.62 & 0.8178 & 8.80    \\ \hline
\multirow{2}{*}{$\text{ResNet50 \cite{resnet}}$} 
    & $\text{ResNet50}_t$ & 370.47 & 0.7952 & -       \\ 
    & + GAT            & 356.91 & 0.9478 & 19.40   \\ \hline
\multirow{2}{*}{$\text{InceptionV3 \cite{inception}}$} 
    & $\text{InceptionV3}_t$ & 350.00 & 0.7105 & -       \\ 
    & + GAT            & 322.00 & 0.8707 & 24.40   \\ \hline
\multirow{2}{*}{$\text{DenseNet121 \cite{densenet}}$} 
    & $\text{DenseNet121}_t$ & 255.77 & 0.6953 & -       \\ 
    & + GAT            & 233.33 & 0.9247 & 33.40   \\ \hline
\multirow{2}{*}{$\text{DenseNet201 \cite{densenet}}$} 
    & $\text{DenseNet201}_t$ & 243.81 & 0.7155 & -       \\ 
    & + GAT            & 225.51 & 0.9214 & 28.00   \\ \hline
\multirow{2}{*}{$\text{GoogLeNet \cite{googlenet}}$} 
    & $\text{GoogLeNet}_t$ & 471.78 & 0.7274 & -       \\ 
    & + GAT            & 311.01 & 0.8631 & 18.80   \\ \hline
\multirow{2}{*}{$\text{ShuffleNetV2 \cite{shufflenet}}$} 
    & $\text{ShuffleNetV2}_t$ & 480.88 & 0.3532 & -       \\ 
    & + GAT            & 333.35 & 0.4598 & 29.70   \\ \hline
\multirow{2}{*}{$\text{MobileNetV2 \cite{23}}$} 
    & $\text{MobileNetV2}_t$ & 490.25 & 0.4525 & -       \\ 
    & + GAT            & 309.52 & 0.4965 & 9.60    \\ \hline
\multirow{2}{*}{$\text{SqueezeNet \cite{squeezenet}}$} 
    & $\text{SqueezeNet}_t$ & 475.32 & 0.7172 & -       \\ 
    & + GAT            & 368.09 & 0.7843 & 9.70    \\ \hline
\multirow{2}{*}{$\text{EfficientNetV2 \cite{effnet}}$} 
    & $\text{EfficientNetV2}_t$ & 355.27 & 0.4352 & -       \\ 
    & + GAT (Proposed SAG-ViT) & 376.43 & 0.9861 & 127.00  \\ \hline
\end{tabular}
}
\end{table}

\textbf{PlantVillage } On PlantVillage (Table~\ref{tab:plantvillage}), SAG-ViT attains an F1 score of 0.9772, representing a +240.00\% improvement over the EfficientNetV2$\rightarrow$ViT baseline. Compared to DenseNet201$\rightarrow$ViT + GAT (F1 = 0.8725), SAG-ViT improves performance by +12.01\%. It also surpasses ResNet50$\rightarrow$ViT + GAT (F1 = 0.8905) by +9.73\%. With a throughput of 371.92 images/s, SAG-ViT remains competitive with lightweight models such as ShuffleNetV2$\rightarrow$ViT + GAT (441.22 images/s) while significantly outperforming them in F1 score (+206.43\%).

\textbf{NCT-CRC-HE-100K } On the NCT-CRC-HE-100K dataset (Table~\ref{tab:nctcrc}), SAG-ViT attains an F1 score of 0.9861, outperforming DenseNet121$\rightarrow$ViT + GAT (F1 = 0.9247) by +6.65\% and ResNet50$\rightarrow$ViT + GAT (F1 = 0.9478) by +4.04\%. SAG-ViT also demonstrates a high throughput of 376.43 images/s, comparable to EfficientNetV2$\rightarrow$ViT (355.27 images/s). Lightweight models such as MobileNetV2$\rightarrow$ViT + GAT exhibit significantly lower absolute F1 scores (F1 = 0.4965), indicating SAG-ViT's competitive performance while maintaining computational efficiency.

\textbf{NWPU-RESISC45 } Table~\ref{tab:nwpuresisc45} highlights the consistent performance of SAG-ViT on NWPU-RESISC45, with an F1 score of 0.9549, marking a +4081.00\% improvement over EfficientNetV2$\rightarrow$ViT. Compared to ResNet50$\rightarrow$ViT + GAT (F1 = 0.9103), SAG-ViT achieves a +4.90\% improvement, while outperforming DenseNet201$\rightarrow$ViT + GAT (F1 = 0.4493) by +112.64\%. Parallel to this significant performance, SAG-ViT maintains a throughput of 365.56 images/s, comparable to other high-performing models such as EfficientNetV2$\rightarrow$ViT (354.76 images/s).

\textbf{UTD } On the UTD (Table~\ref{tab:utd}), SAG-ViT yields an F1 score of 0.9615, surpassing EfficientNetV2$\rightarrow$ViT by +94.40\% and ResNet50$\rightarrow$ViT + GAT (F1 = 0.8453) by +13.76\%. DenseNet121$\rightarrow$ViT + GAT attains an F1 score of 0.8377, which SAG-ViT improves upon by +14.77\%. SAG-ViT also maintains efficient throughput (254.18 images/s) while outperforming lightweight models such as ShuffleNetV2$\rightarrow$ViT + GAT (324.27 images/s) in F1 score by +40.94\%.


\begin{table}[h!]
\centering
\caption{Performance comparison on the NWPU-RESISC45 dataset.}
\label{tab:nwpuresisc45}
\resizebox{\columnwidth}{!}{
\begin{tabular}{l|lccc}
\hline
\textbf{Model} & \textbf{Method} & \textbf{(\(\mathcal{T}\)) (images/s)} & \textbf{\(F_1\)} & \textbf{\(\Delta F_1\) (\%)} \\ \hline
\multirow{2}{*}{$\text{VGG16 \cite{vgg}}$} 
    & $\text{VGG16}_t$ & 263.22 & 0.0039 & -       \\ 
    & + GAT            & 264.43 & 0.4114 & 10448.70 \\ \hline
\multirow{2}{*}{$\text{VGG19 \cite{vgg}}$} 
    & $\text{VGG19}_t$ & 226.12 & 0.0454 & -       \\ 
    & + GAT            & 229.66 & 0.3844 & 747.80  \\ \hline
\multirow{2}{*}{$\text{ResNet50 \cite{resnet}}$} 
    & $\text{ResNet50}_t$ & 402.65 & 0.0472 & -       \\ 
    & + GAT            & 375.83 & 0.9103 & 1830.30 \\ \hline
\multirow{2}{*}{$\text{InceptionV3 \cite{inception}}$} 
    & $\text{InceptionV3}_t$ & 350.00 & 0.0701 & -       \\ 
    & + GAT            & 340.00 & 0.8707 & 1144.10 \\ \hline
\multirow{2}{*}{$\text{DenseNet121 \cite{densenet}}$} 
    & $\text{DenseNet121}_t$ & 363.32 & 0.0348 & -       \\ 
    & + GAT            & 265.62 & 0.4381 & 1159.20 \\ \hline
\multirow{2}{*}{$\text{DenseNet201 \cite{densenet}}$} 
    & $\text{DenseNet201}_t$ & 242.51 & 0.0364 & -       \\ 
    & + GAT            & 172.23 & 0.4493 & 1133.30 \\ \hline
\multirow{2}{*}{$\text{GoogLeNet \cite{googlenet}}$} 
    & $\text{GoogLeNet}_t$ & 630.99 & 0.0365 & -       \\ 
    & + GAT            & 410.58 & 0.3720 & 919.20  \\ \hline
\multirow{2}{*}{$\text{ShuffleNetV2 \cite{shufflenet}}$} 
    & $\text{ShuffleNetV2}_t$ & 698.57 & 0.0182 & -       \\ 
    & + GAT            & 431.78 & 0.1808 & 893.40  \\ \hline
\multirow{2}{*}{$\text{MobileNetV2 \cite{23}}$} 
    & $\text{MobileNetV2}_t$ & 625.36 & 0.0203 & -       \\ 
    & + GAT            & 432.82 & 0.1667 & 720.30  \\ \hline
\multirow{2}{*}{$\text{SqueezeNet \cite{squeezenet}}$} 
    & $\text{SqueezeNet}_t$ & 622.38 & 0.0430 & -       \\ 
    & + GAT            & 487.92 & 0.3913 & 809.30  \\ \hline
\multirow{2}{*}{$\text{EfficientNetV2 \cite{effnet}}$} 
    & $\text{EfficientNetV2}_t$ & 354.76 & 0.0228 & -       \\ 
    & + GAT (Proposed SAG-ViT) & 365.56 & 0.9549 & 4081.00  \\ \hline
\end{tabular}
}
\end{table}

\begin{table}[h!]
\centering
\caption{Performance comparison on the UTD.}
\label{tab:utd}
\resizebox{\columnwidth}{!}{
\begin{tabular}{l|lccc}
\hline
\textbf{Model} & \textbf{Method} & \textbf{(\(\mathcal{T}\)) (images/s)} & \textbf{\(F_1\)} & \textbf{\(\Delta F_1\) (\%)} \\ \hline
\multirow{2}{*}{$\text{VGG16 \cite{vgg}}$} 
    & $\text{VGG16}_t$ & 215.88 & 0.2947 & -       \\ 
    & + GAT            & 229.63 & 0.7973 & 170.00  \\ \hline
\multirow{2}{*}{$\text{VGG19 \cite{vgg}}$} 
    & $\text{VGG19}_t$ & 198.48 & 0.2949 & -       \\ 
    & + GAT            & 209.28 & 0.8159 & 176.00  \\ \hline
\multirow{2}{*}{$\text{ResNet50 \cite{resnet}}$} 
    & $\text{ResNet50}_t$ & 388.07 & 0.7823 & -       \\ 
    & + GAT            & 352.83 & 0.8453 & 8.50    \\ \hline
\multirow{2}{*}{$\text{InceptionV3 \cite{inception}}$} 
    & $\text{InceptionV3}_t$ & 350.00 & 0.7012 & -       \\ 
    & + GAT            & 340.00 & 0.8707 & 24.30   \\ \hline
\multirow{2}{*}{$\text{DenseNet121 \cite{densenet}}$} 
    & $\text{DenseNet121}_t$ & 232.19 & 0.5797 & -       \\ 
    & + GAT            & 242.54 & 0.8377 & 44.90   \\ \hline
\multirow{2}{*}{$\text{DenseNet201 \cite{densenet}}$} 
    & $\text{DenseNet201}_t$ & 167.04 & 0.6823 & -       \\ 
    & + GAT            & 178.25 & 0.8742 & 28.30   \\ \hline
\multirow{2}{*}{$\text{GoogLeNet \cite{googlenet}}$} 
    & $\text{GoogLeNet}_t$ & 290.78 & 0.4679 & -       \\ 
    & + GAT            & 311.44 & 0.8134 & 73.20   \\ \hline
\multirow{2}{*}{$\text{ShuffleNetV2 \cite{shufflenet}}$} 
    & $\text{ShuffleNetV2}_t$ & 294.94 & 0.5621 & -       \\ 
    & + GAT            & 324.27 & 0.6829 & 21.80   \\ \hline
\multirow{2}{*}{$\text{MobileNetV2 \cite{23}}$} 
    & $\text{MobileNetV2}_t$ & 290.95 & 0.5739 & -       \\ 
    & + GAT            & 328.39 & 0.6602 & 15.50   \\ \hline
\multirow{2}{*}{$\text{SqueezeNet \cite{squeezenet}}$} 
    & $\text{SqueezeNet}_t$ & 279.02 & 0.7332 & -       \\ 
    & + GAT            & 312.24 & 0.7677 & 4.90    \\ \hline
\multirow{2}{*}{$\text{EfficientNetV2 \cite{effnet}}$} 
    & $\text{EfficientNetV2}_t$ & 244.39 & 0.4938 & -       \\ 
    & + GAT (Proposed SAG-ViT) & 254.18 & 0.9615 & 94.40   \\ \hline
\end{tabular}
}
\end{table}

\subsection{Loss Landscape Analysis}

Figure \ref{fig:landscapes} visualizes the loss \( \mathcal{L} \) as a function of perturbations along two random directions \( \mathbf{d}_1 \) and \( \mathbf{d}_2 \) in the parameter space for key layers of the proposed model, including the EfficientNetV2 backbone's final convolutional layer, the final graph-based layer, and the final Transformer layer. By visualizing the plotted surfaces for each layer, we understand how well our model is learning and optimizing.

\begin{table*}[h!]
    \centering
    \caption{Hardware RAM and GPU (VRAM) consumption \textbf{(\%)} of different [CNN(\(\rightarrow\))ViT+GAT] and Transformer architectures on benchmark datasets. The minimum values in each column are highlighted in \textcolor{blue}{blue} and the second minimum values in \textcolor{green}{green}.}
    \label{tab:hardware_consumption}
    \resizebox{\textwidth}{!}{\renewcommand{\arraystretch}{1} 
\begin{tabular}{@{}lcccccccccccc@{}} 
        \toprule
        \multirow{2}{*}{\textbf{Method}}& \multicolumn{2}{c}{\textbf{CIFAR-10}} & \multicolumn{2}{c}{\textbf{GTSRB}} & \multicolumn{2}{c}{\textbf{NCT-CRC-HE-100K}} & \multicolumn{2}{c}{\textbf{PlantVillage}} & \multicolumn{2}{c}{\textbf{NWPU-RESISC45}} & \multicolumn{2}{c}{\textbf{UTD}} \\
        \cmidrule(lr){2-3} \cmidrule(lr){4-5} \cmidrule(lr){6-7} \cmidrule(lr){8-9} \cmidrule(lr){10-11} \cmidrule(lr){12-13}
        & \textbf{RAM} & \textbf{GPU} & \textbf{RAM} & \textbf{GPU} & \textbf{RAM} & \textbf{GPU} & \textbf{RAM} & \textbf{GPU} & \textbf{RAM} & \textbf{GPU} & \textbf{RAM} & \textbf{GPU} \\
        \midrule
        ViT - S (Not Backbone)& 11.37 & 30.62 & 12.30 & 29.70 & 12.10 & 24.28 & 12.50 & 27.32 & 13.20 & 29.54 & 17.80 & \textcolor{green}{11.17}\\
        ViT - L (Not Backbone)& 15.17 & 81.87 & 13.60 & 33.72 & 14.80 & 33.21 & 13.50 & 35.98 & 13.90 & 34.96 & 21.70 & 17.41 \\
        DenseNet201 & 9.70 & 24.32 & 11.00 & 19.15 & 11.10 & 19.14 & 11.30 & \textcolor{green}{17.97} & 11.00 & 18.77 & 12.60 & 12.16 \\
        VGG16 & 11.50 & 42.19 & 11.30 & 29.76 & 12.10 & 29.77 & 11.60 & 30.58 & 11.10 & 29.27 & 13.10 & 17.72 \\
        VGG19 & 11.80 & 55.93 & 11.10 & 29.27 & 11.80 & 29.27 & 11.40 & 30.24 & 11.20 & 29.77 & 12.80 & 21.56 \\
        DenseNet121 & \textcolor{green}{8.10} & 28.80 & 10.90 & 16.81 & 11.20 & 16.81 & \textcolor{green}{11.20} & 17.52 & \textcolor{green}{10.80} & 16.88 & 12.60 & 10.54 \\
        InceptionV3 & 10.20 & 29.40 & 10.50 & 22.10 & 11.40 & 24.70 & 11.50 & 24.20 & \textcolor{green}{10.80} & 20.32 & \textcolor{green}{12.10}& 11.32 \\
        ResNet50 & 9.50 & 23.10 & 10.70 & 18.90 & 11.10 & 20.30 & \textcolor{green}{11.20} & 21.20 & 10.90 & 18.92 & 12.70 & 13.51 \\
        MobileNetV2 & 12.41 & 22.50 & 11.00 & 21.63 & \textcolor{green}{7.60} & \textcolor{green}{14.08} & 11.30 & 19.39 & 10.90 & 21.63 & 18.10 & 14.42 \\
        ShuffleNet & 9.80 & \textcolor{blue}{12.13} & 10.80 & \textcolor{blue}{15.14} & 8.70 & 11.50 & 11.30 & \textcolor{blue}{16.06} & 10.90 & 15.10 & 13.30 & \textcolor{green}{11.11}\\
        SqueezeNet & 11.72 & 19.37 & \textcolor{blue}{7.50} & \textcolor{green}{10.34} & \textcolor{blue}{7.70} & \textcolor{blue}{10.62} & 11.40 & \textcolor{blue}{16.02} & 11.00 & 15.21 & 13.10 & 11.21 \\
        GoogLeNet & \textcolor{green}{8.10} & 19.06 & 10.70 & \textcolor{blue}{15.07} & 9.40 & 16.12 & 11.30 & \textcolor{blue}{16.05} & \textcolor{green}{10.80} & \textcolor{green}{15.08} & 25.50 & 13.37 \\
        SAG-ViT (Proposed) & \textcolor{blue}{7.24} & 33.12 & \textcolor{green}{9.20} & 36.38 & \textcolor{green}{7.60} & 37.32 & \textcolor{blue}{8.20} & 39.32 & \textcolor{blue}{10.72} & \textcolor{blue}{11.62} & \textcolor{blue}{7.20}& \textcolor{blue}{10.21}\\
        \bottomrule
\end{tabular}
}
 
\end{table*}

As shown in Figure \ref{fig:landsape1}, the final backbone layer exhibits a well-conditioned loss landscape. The gentle slope around the global minimum ensures stable convergence, while the steeper gradients \( \|\nabla \mathcal{L}(\mathbf{w})\| \) outside the basin ensure that optimization rapidly approaches the minima. In Figure \ref{fig:landsape2}, the GAT's loss landscape displays a relatively smooth and convex shape with a clearly defined global minimum. This smoothness ensures stable convergence during gradient descent, as the gradient vector \( \nabla \mathcal{L}(\mathbf{w}) = \frac{\partial \mathcal{L}}{\partial \mathbf{w}} \) is well-behaved and avoids chaotic updates. The basin-shaped structure implies that the weight space around the solution has low curvature, leading to smaller eigenvalues in the Hessian matrix \( \mathbf{H} = \nabla^2 \mathcal{L}(\mathbf{w}) \), which correlates to optimal generalization properties. The Transformer's loss landscape, illustrated in Figure \ref{fig:landsape3}, is characterized by a low-curvature region around the global minimum. This flatness indicates that the eigenvalues \( \lambda_i \) of \( \mathbf{H} \) near the minimum are close to zero, depicting the model's generalizing capabilities. Moreover, the symmetry of the valley in all directions \( \mathbf{d}_1, \mathbf{d}_2 \) indicates a well-conditioned parameter space and is less susceptible to overfitting. These loss landscapes exhibit our model's effective training, with flat minima and smooth surfaces indicating good generalization.

\subsection{Hardware Efficiency}

We also evaluated the hardware efficiency of SAG-ViT against other [CNN(\(\rightarrow\))ViT+GAT] architectures in terms of RAM and GPU VRAM usage. Table~\ref{tab:hardware_consumption} details the resource consumption of each architecture across the different datasets. SAG-ViT achieves a favorable balance between resource consumption and computational complexity due to the efficient integration of CNN-based feature patching and GAT into a Transformer architecture. Compared to standard Transformers like ViT-S and ViT-L, our method provides a more resource-efficient alternative while leveraging the advantages of graph-based attention to achieve better feature representation and classification performance.

\begin{table}[h!]
\centering
\caption{Hardware utilization comparison of [CNN(\(\rightarrow\))ViT] architectures with and without the GAT module. GFLOPs (\(\mathbb{G}\)) and parameters in millions (\(\theta\)) are reported for both configurations, along with their percentage reductions (\(\downarrow\)), calculated relative to the version without GAT.}
\label{tab:model_comparison}
\resizebox{\columnwidth}{!}{\small
\begin{tabular}{lp{1.2cm}p{1.2cm}p{1cm}p{1.2cm}}
\hline
\textbf{Method} & \textbf{\(\mathbb{G}\)} & \textbf{\(\mathbb{G} \downarrow (\%)\)} & \textbf{\(\theta (M)\)} & \textbf{\(\theta \downarrow (\%)\)} \\ \hline
$\text{VGG16}_t$               & 20.08  & -       & 16.97  & -       \\ 
+ GAT               & 20.04  & 0.19    & 14.73  & 13.2    \\ 
$\text{VGG19}_t$               & 25.52  & -       & 22.28  & -       \\ 
+ GAT               & 25.48  & 0.15    & 20.04  & 10.05   \\ 
$\text{ResNet50}_t$            & 5.45   & -       & 26.42  & -       \\ 
+ GAT               & 5.39   & 1.08    & 22.64  & 14.3    \\ 
$\text{InceptionV3}_t$         & 2.86   & -       & 23.58  & -       \\ 
+ GAT               & 2.84   & 0.46    & 19.38  & 17.81   \\ 
$\text{DenseNet121}_t$         & 3.83   & -       & 09.43   & -       \\ 
+ GAT               & 3.78   & 1.15    & 06.97   & 26.1    \\ 
$\text{DenseNet201}_t$         & 5.78   & -       & 20.95  & -       \\ 
+ GAT               & 5.73   & 0.91    & 18.10  & 13.59   \\ 
$\text{GoogLeNet}_t$           & 2.02   & -       & 08.07   & -       \\ 
+ GAT               & 1.97   & 2.17    & 05.61   & 30.48   \\ 
$\text{ShuffleNetV2}_t$ & 0.24 & -       & 03.73   & -       \\ 
+ GAT               & 0.20   & 18.11   & 01.26   & 66.05   \\ 
$\text{MobileNetV2}_t$       & 0.47   & -       & 04.81   & -       \\ 
+ GAT               & 0.43   & 9.80    & 02.24   & 53.50   \\ 
$\text{SqueezeNet}_t$      & 1.00   & -       & 02.99   & -       \\ 
+ GAT               & 0.97   & 3.87    & 00.75   & 74.99   \\ 
$\text{EfficientNetV2}_t$   & 3.83   & -       & 09.76   & -       \\ 
+ GAT (Proposed SAG-ViT)        & 2.98   & 22.38   & 06.39   & 34.52   \\ \hline
\end{tabular}
}
\end{table}

\begin{table*}[t]
    \centering
    \caption{Ablation Study Results on Benchmark Datasets}
    \label{tab:ablation_study}
    \resizebox{\textwidth}{!}{\begin{tabular}{l|cccc|cccc}
        \multirow{2}{*}{Setup} & \multicolumn{4}{c|}{CIFAR-10} & \multicolumn{4}{c}{NCT-CRC-HE-100K} \\
        \cline{2-9}
         & $F_1$ & $RAM_\text{GB}$ & $VRAM_\text{GB}$ & $Time_\text{epoch}$ & $F_1$ & $RAM_\text{GB}$ & $VRAM_\text{GB}$ & $Time_\text{epoch}$ \\
        \hline
        Backbone + GAT (No Transformer) & 0.779 & 5.601 & 4.902 & 00:14:30 & 0.956 & 4.801 & 3.504 & 00:14:13 \\
        Backbone + Transformer (No GAT) & 0.759 & 3.102 & 4.501 & 00:16:07 & 0.169 & 17.504 & 4.702 & 00:13:17 \\
        GAT + Transformer (No Backbone) & 0.503 & 4.304 & 5.302 & 01:33:24 & 0.657 & 18.793 & 1.591 & 02:36:35 \\
        \hline
        \multirow{2}{*}{Setup} & \multicolumn{4}{c|}{PlantVillage} & \multicolumn{4}{c}{NWPU-RESISC45} \\
        \cline{2-9}
         & $F_1$ & $RAM_\text{GB}$ & $VRAM_\text{GB}$ & $Time_\text{epoch}$ & $F_1$ & $RAM_\text{GB}$ & $VRAM_\text{GB}$ & $Time_\text{epoch}$ \\
        \hline
        Backbone + GAT (No Transformer) & 0.899 & 2.401 & 3.402 & 00:02:27 & 0.789 & 3.304 & 3.401 & 00:03:20 \\
        Backbone + Transformer (No GAT) & 0.033 & 2.802 & 3.801 & 00:02:35 & 0.007 & 4.101 & 4.504 & 00:03:25 \\
        GAT + Transformer (No Backbone) & 0.785 & 4.687 & 2.512 & 00:30:16 & 0.526 & 4.522 & 3.927 & 01:25:10 \\
        \hline
        \multirow{2}{*}{Setup} & \multicolumn{4}{c|}{GTSRB} & \multicolumn{4}{c}{UTD} \\
        \cline{2-9}
         & $F_1$ & $RAM_\text{GB}$ & $VRAM_\text{GB}$ & $Time_\text{epoch}$ & $F_1$ & $RAM_\text{GB}$ & $VRAM_\text{GB}$ & $Time_\text{epoch}$ \\
        \hline
        Backbone + GAT (No Transformer) & 0.966 & 6.102 & 6.402 & 00:05:03 & 0.927 & 3.402 & 14.801 & 00:01:24 \\
        Backbone + Transformer (No GAT) & 0.009 & 3.804 & 3.204 & 00:05:20 & 0.390 & 2.902 & 13.705 & 00:01:22 \\
        GAT + Transformer (No Backbone) & 0.048 & 5.405 & 2.502 & 00:55:13 & 0.532 & 3.502 & 14.321 & 00:25:32 \\
    \end{tabular}}
 
\end{table*}

On CIFAR-10, we find that our model records the lowest RAM consumption at 7.24\%, outperforming models like VGG16 (11.5\%) and DenseNet201 (9.7\%). Its GPU VRAM usage of 33.12\% remains moderate, especially when compared to ViT-L, which requires 81.87\% GPU VRAM. This efficiency reflects the balance achieved by integrating EfficientNetV2 with sparse \textit{k}-connectivity graphs. For GTSRB, our model utilizes 9.20\% RAM, making it the second most efficient model after SqueezeNet (7.50\%). While the GPU VRAM usage of 36.38\% is higher than that of lightweight models like ShuffleNet (15.14\%), it remains significantly lower than ViT-L’s 33.72\%. On the NCT-CRC-HE-100K dataset, we observe that our model achieves the second-lowest RAM consumption at 7.60\%, close to SqueezeNet’s 7.70\%. However, the GPU VRAM usage of 37.32\% is higher compared to models like DenseNet121 (16.81\%) but remains reasonable given the additional components of our architecture.

For PlantVillage, our model registers the lowest RAM consumption at 8.20\%, outperforming MobileNetV2 and GoogLeNet (both 11.30\%). The GPU VRAM usage of 39.32\% is higher than that of some [CNN(\(\rightarrow\))ViT] architectures but reflects the computational requirements of the GAT and Transformer components. On the NWPU-RESISC45 dataset, we find that our model has the lowest RAM usage at 10.72\%, followed closely by GoogLeNet (10.80\%). Additionally, it records the lowest GPU VRAM consumption at 11.62\%, underscoring its efficiency on this dataset. Finally, for the UTD dataset, our records the lowest RAM usage (7.20\%) and GPU VRAM usage (10.21\%) among all tested configurations, including lightweight models like SqueezeNet and DenseNet121.

\subsection{Ablation Study}

\begin{figure*}[h!]
    \centering
    \begin{subfigure}[b]{0.32\textwidth}
        \centering
        \includegraphics[width=\textwidth]{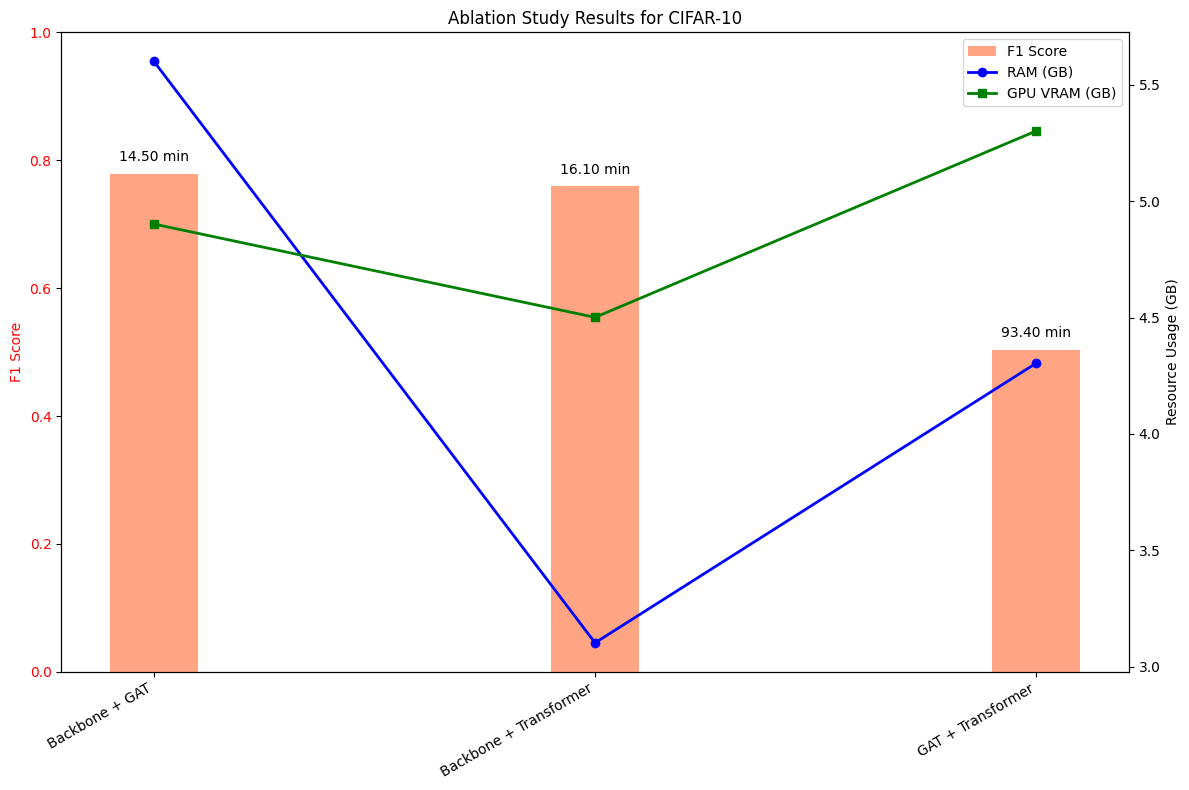}
        \caption{Ablation results for CIFAR-10.}
        \label{fig:ablationCifar}
    \end{subfigure}
    \hfill
    \begin{subfigure}[b]{0.32\textwidth}
        \centering
        \includegraphics[width=\textwidth]{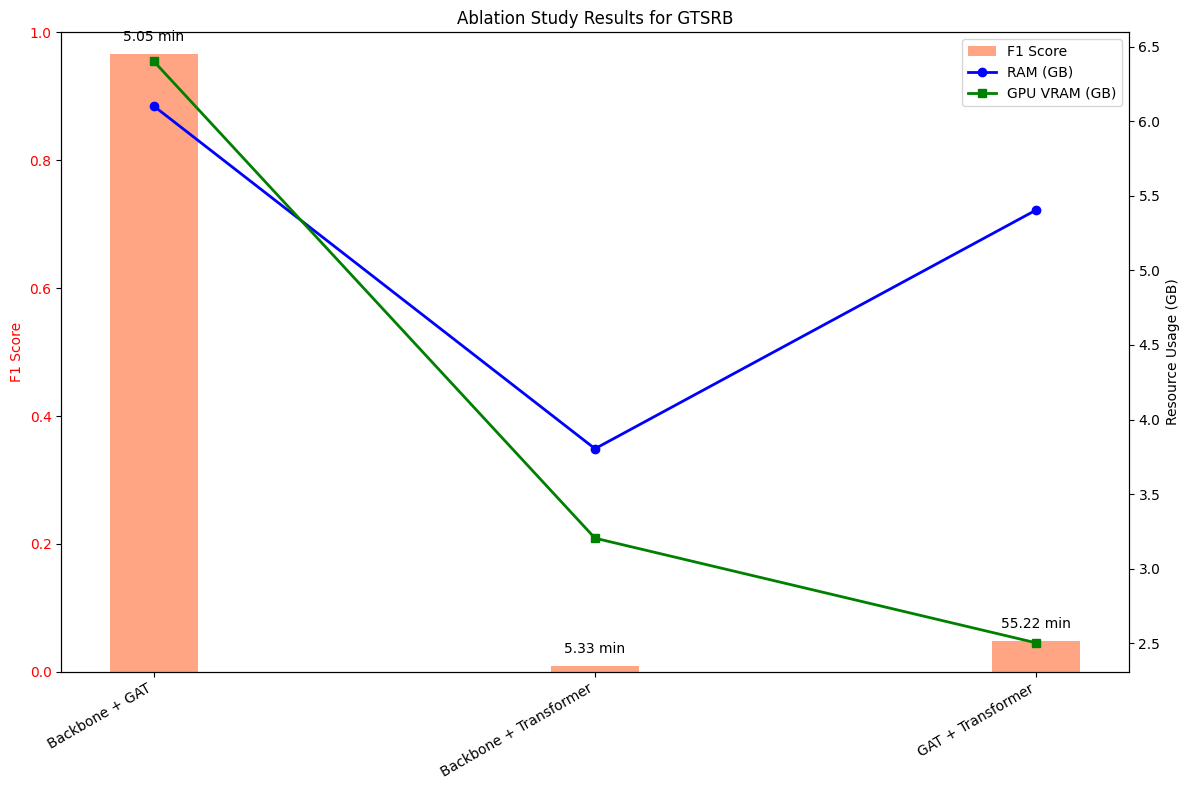}
        \caption{Ablation results for GTSRB.}
        \label{fig:ablationGTSRB}
    \end{subfigure}
    \hfill
    \begin{subfigure}[b]{0.32\textwidth}
        \centering
        \includegraphics[width=\textwidth]{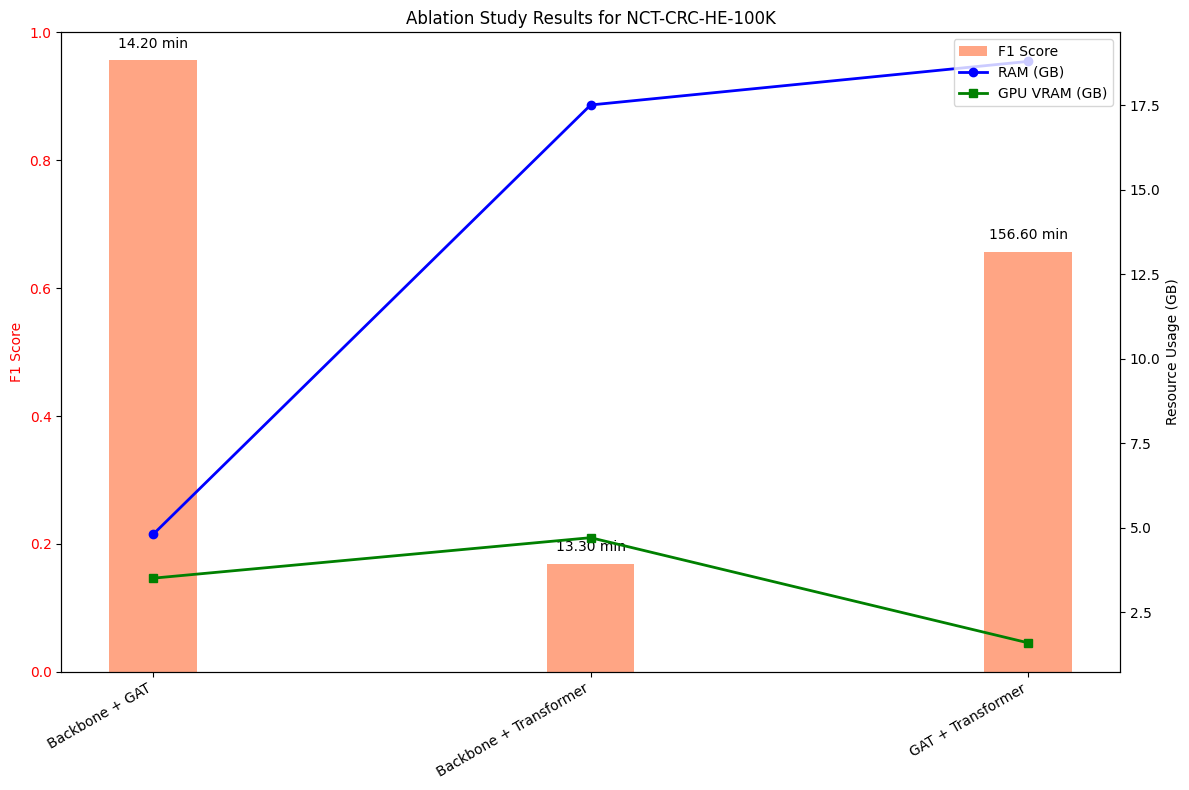}
        \caption{Ablation results for NCT-CRC-HE-100K.}
        \label{fig:ablationNCT}
    \end{subfigure}
    \\
    \begin{subfigure}[b]{0.32\textwidth}
        \centering
        \includegraphics[width=\textwidth]{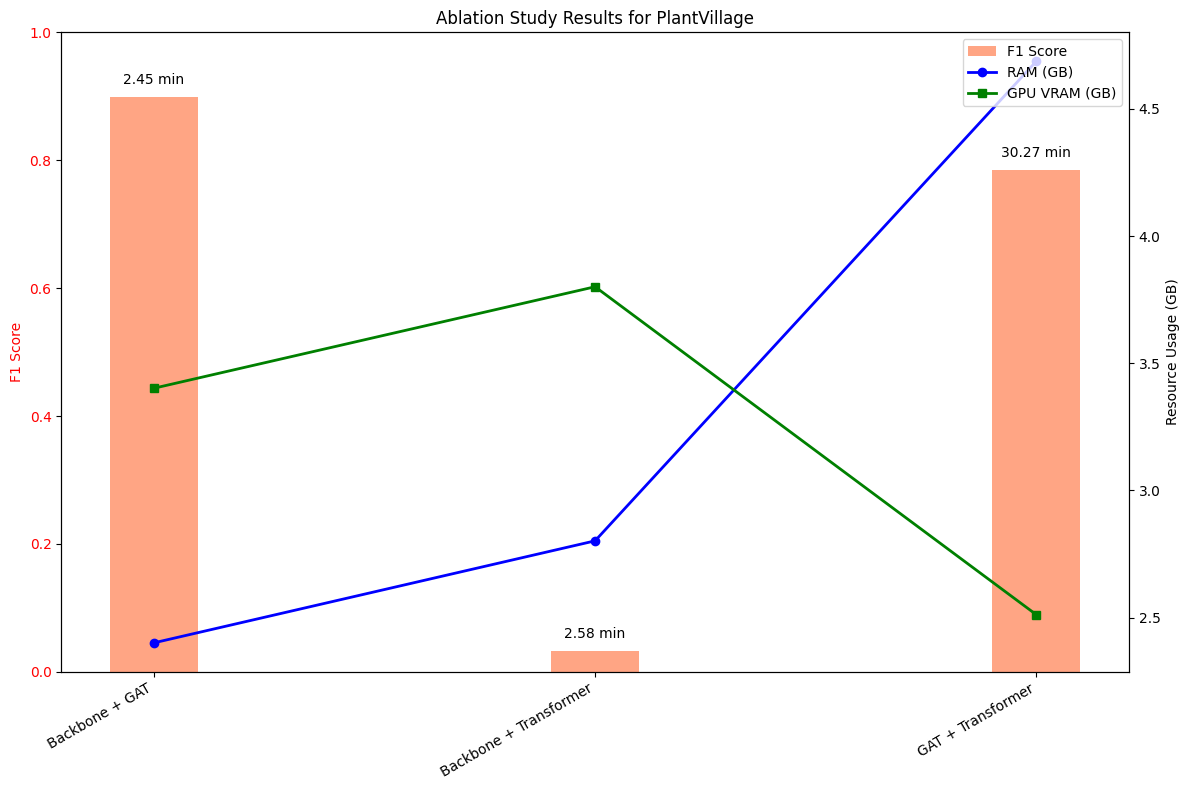}
        \caption{Ablation results for PlantVillage.}
        \label{fig:ablationPlant}
    \end{subfigure}
    \hfill
    \begin{subfigure}[b]{0.32\textwidth}
        \centering
        \includegraphics[width=\textwidth]{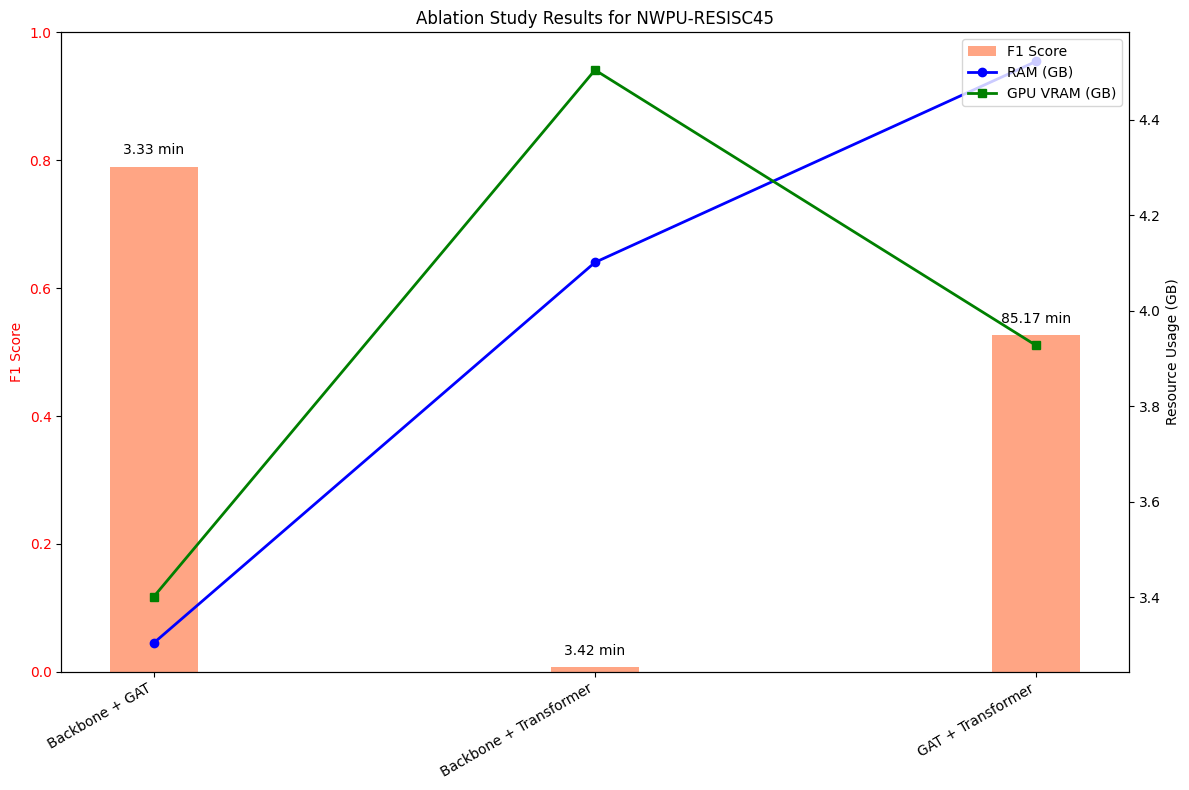}
        \caption{Ablation results for NWPU-RESISC45.}
        \label{fig:ablationNWPU}
    \end{subfigure}
    \hfill
    \begin{subfigure}[b]{0.32\textwidth}
        \centering
        \includegraphics[width=\textwidth]{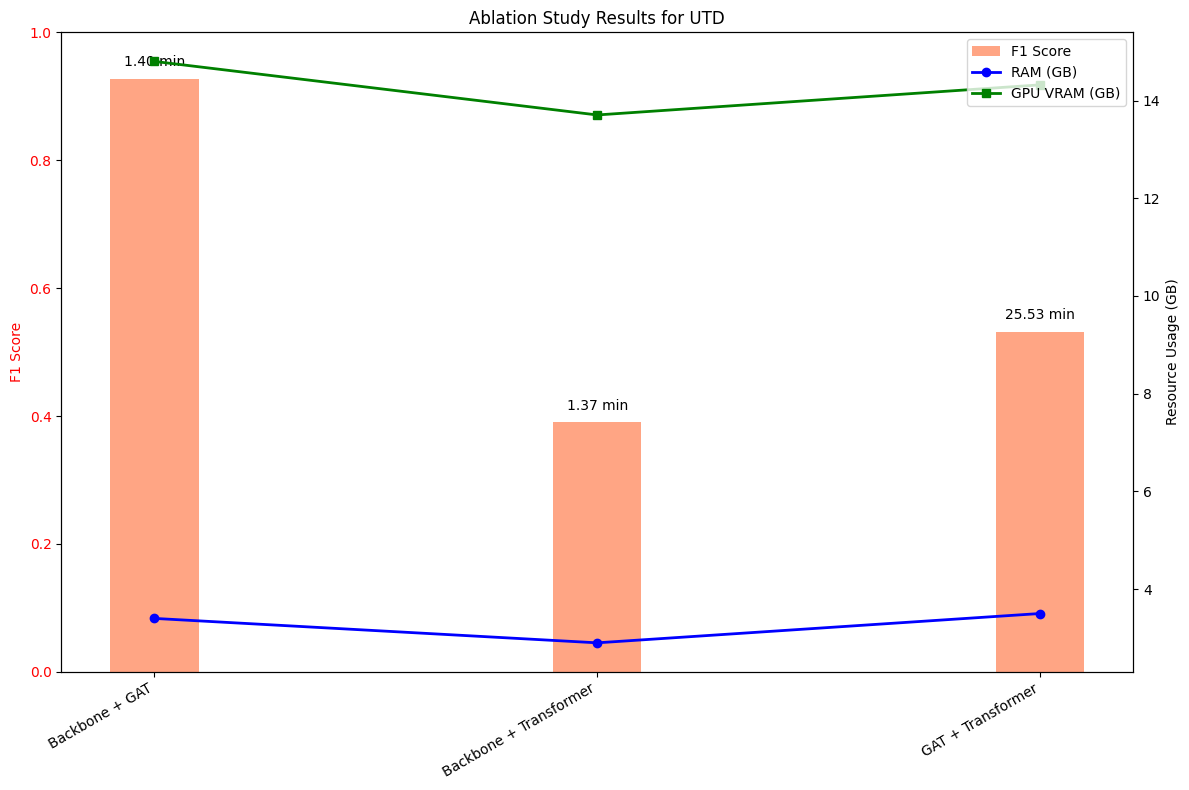}
        \caption{Ablation results for UTD.}
        \label{fig:ablationUTD}
    \end{subfigure}
    \caption{Ablation study results for CIFAR-10, GTSRB, NCT-CRC-HE-100K, NWPU-RESISC45, PlantVillage, and UTD datasets.}
    \label{fig:ablationResults}
\end{figure*}

In this section, we conduct an ablation study to evaluate the contribution of key components in SAG-ViT: the EfficientNetV2 backbone, the Graph Attention Network (GAT), and the Transformer encoder. The results are summarized in Table ~\ref{tab:ablation_study}. Ablation experiments were conducted on all 6 datasets and involved the following configurations:


\begin{enumerate}
    \item \textbf{Backbone + GAT (Without Transformer):}
        To assess the role of the Transformer in capturing global dependencies, we exclude the Transformer encoder. In this setup, the EfficientNetV2 backbone extracts feature embeddings, which are processed by the GAT to refine local relationships. The class predictions are generated directly from the aggregated node representations produced by the GAT.

    \item \textbf{Backbone + Transformer (Without GAT):}  
        To evaluate the importance of the GAT in modeling local dependencies and refining node features, we omit it from the architecture. The feature embeddings extracted by the backbone are fed directly into the Transformer encoder, which is tasked with learning both local and global relationships without the localized attention mechanism provided by the GAT.
        
    \item \textbf{GAT + Transformer (Without Backbone):}  
        In this configuration, we omit the EfficientNetV2 backbone to evaluate the impact of multi-scale feature extraction and feature map-based patching. Instead of using backbone-generated feature maps, graphs are constructed directly on the input image by patching the original image with a 4x4 patch size. Each patch is treated as a node, and edges are defined using weighted \textit{k}-connectivity. This setup allows us to analyze the effectiveness and efficiency of feature map-based patching in capturing rich semantic information and spatial relationships, in terms of both training time and quantitative performance. 

\end{enumerate}

\textbf{Settings } Training was conducted under the same conditions as our other experiments, with additional logging of training time and resource utilization (GPU VRAM and RAM) to analyze computational efficiency. A batch size of 128 was used for configurations 1 and 2, but it was reduced to 32 for the 3\textsuperscript{rd} configuration due to GPU memory limitations caused by excessive patching and graph construction directly on the raw image. The Adam optimizer \cite{AdamOpt} was employed with a learning rate of 0.001 and a cosine decay schedule, including a 10-epoch linear warm-up phase. For all configurations, the input image resolution was standardized to 224\textsuperscript{2}, gradient clipping with a maximum norm of 1 was applied, and model performance was evaluated using F1 score, throughput (images/s), and training time per epoch as metrics.

\textbf{CIFAR-10 } Figure~\ref{fig:ablationCifar} shows that the Backbone + GAT configuration attains an F1 score of 0.779. The RAM and GPU VRAM usage remains moderate at 5.601 GB and 4.902 GB, respectively, and the training time per epoch is 14:30 minutes. The Backbone + Transformer configuration achieves a slightly lower F1 score of 0.759, showing the Transformer's capability to model global relationships but without the local refinement provided by the GAT. This setup uses the least RAM (3.102 GB) but has a slightly higher VRAM usage (4.501 GB), with a training time of 16:07 minutes. In contrast, the GAT + Transformer setup, which bypasses the EfficientNetV2 backbone, results in a drastic drop in F1 score to 0.503, with significantly higher GPU VRAM usage (5.302 GB) and training time of 1:33:24 hours. These results indicate the role of the backbone in providing high-quality feature maps, reducing training time, and ensuring computational efficiency.

\textbf{NCT-CRC-HE-100K } Figure~\ref{fig:ablationNCT} illustrates that the Backbone + GAT configuration yields a strong F1 score of 0.956, with moderate resource consumption (4.801 GB RAM, 3.504 GB GPU VRAM) and a training time of 14:13 minutes per epoch. Without the GAT, the Backbone + Transformer configuration suffers a severe performance drop, achieving an F1 score of only 0.169. This setup demands significantly higher RAM usage (17.504 GB) and a relatively higher GPU VRAM (4.702 GB), with a reduced training time of 13:17 minutes. The GAT + Transformer configuration attains an F1 score of 0.657, reflecting the importance of multi-scale feature extraction by the backbone. This setup consumes the highest RAM (18.793 GB), with reduced GPU VRAM usage (1.591 GB) but an extremely long training time of 2:36:35 hours.

\textbf{PlantVillage } On PlantVillage, Figure~\ref{fig:ablationPlant} demonstrates that the Backbone + GAT configuration produces an F1 score of 0.899 while maintaining the lowest hardware demands (2.401 GB RAM and 3.402 GB GPU VRAM) with a training time of just 2:27 minutes per epoch. Without the GAT, the Backbone + Transformer setup performs poorly, with an F1 score of only 0.033. This setup shows slightly higher RAM and GPU VRAM usage (2.802 GB and 3.801 GB, respectively) and a marginally higher training time of 2:35 minutes. The GAT + Transformer configuration yields a reasonable F1 score of 0.785, highlighting the backbone's critical role in improving performance. It consumes 4.687 GB RAM, 2.512 GB GPU VRAM, and requires 30:16 minutes per epoch.

\textbf{NWPU-RESISC45 } For NWPU-RESISC45, Figure~\ref{fig:ablationNWPU} shows that the Backbone + GAT configuration produces an F1 score of 0.789 with moderate resource usage (3.304 GB RAM and 3.401 GB GPU VRAM) and a training time of 3:20 minutes. The Backbone + Transformer configuration performs extremely poorly with an F1 score of 0.007, despite consuming higher RAM (4.101 GB) and GPU VRAM (4.504 GB). The GAT + Transformer configuration, bypassing the backbone, yields an F1 score of 0.526, with higher hardware demands (4.522 GB RAM and 3.927 GB GPU VRAM) and an extensive training time of 1:25:10 hours.

\textbf{GTSRB } For the GTSRB dataset, Figure~\ref{fig:ablationGTSRB} highlights that the Backbone + GAT configuration delivers the best performance, with an F1 score of 0.966, moderate RAM usage (6.102 GB), and 6.402 GB GPU VRAM. The training time per epoch is 5:03 minutes. The Backbone + Transformer setup results in a near-zero F1 score (0.009), despite lower hardware requirements (3.804 GB RAM and 3.204 GB GPU VRAM). The GAT + Transformer setup produces a better F1 score of 0.048 but incurs higher GPU VRAM consumption (2.502 GB) and a lengthy training time of 55:13 minutes per epoch.

\textbf{UTD } For the UTD, Figure~\ref{fig:ablationUTD} shows that the Backbone + GAT configuration yields an F1 score of 0.927 while using 3.402 GB RAM and 14.801 GB GPU VRAM, with a training time of 1:24 minutes. The Backbone + Transformer setup produces an F1 score of 0.390, with lower resource demands (2.902 GB RAM and 13.705 GB GPU VRAM). The GAT + Transformer setup, however, achieves a lower F1 score of 0.532, consuming 3.502 GB RAM and 14.321 GB GPU VRAM, with a training time of 25:32 minutes per epoch.


\section{Conclusion and Future Work}
\label{label_conclusion}

This paper introduces SAG-ViT, a novel transformer architecture that combines multi-scale feature extraction, graph-based modeling, and  Transformer-based global dependency learning into a unified framework for image classification. It addresses the limitations of existing approaches by combining the strengths of convolutional backbones, GATs, and Transformers through efficient feature extraction, patch-based graph construction, and the interplay between local and global attention mechanisms. Our results across six diverse datasets validate that incorporating graph-based modeling into ViT architectures not only enhances representation learning but also balances computational efficiency and effectiveness. We hope that SAG-ViT's adaptability to diverse datasets can be leveraged for fine-tuning in domain-specific contexts, such as medical imaging \cite{conc_med1, conc_med2, conc_med3, conc_med4}, especially with the advent of Transformer-based models. We also hope that its hybrid architecture can inspire future research into unified Transformer frameworks and applications across tasks like object detection \cite{conc_od1, conc_od2, conc_od3}, segmentation \cite{conc_seg1, conc_seg2}, and multimodal learning \cite{conc_mm1, conc_mm2, conc_mm3}.

{
    \small
    \bibliographystyle{ieeenat_fullname}
    \bibliography{main}
}


\end{document}